\documentclass{article}

\usepackage{iclr2027_conference,times}

\usepackage{amsmath,amsfonts,bm}

\def\eqref#1{equation~\ref{#1}}

\def\1{\bm{1}}

\DeclareMathAlphabet{\mathsfit}{\encodingdefault}{\sfdefault}{m}{sl}
\SetMathAlphabet{\mathsfit}{bold}{\encodingdefault}{\sfdefault}{bx}{n}

\usepackage[T1]{fontenc}
\usepackage[utf8]{inputenc}
\usepackage{microtype}
\usepackage{hyperref}
\usepackage{url}
\hypersetup{hidelinks}
\usepackage{graphicx}
\usepackage{booktabs}
\usepackage{multirow}
\usepackage{amsmath}
\usepackage{amssymb}
\usepackage{mathtools}
\usepackage{xcolor}
\usepackage{enumitem}
\usepackage{wrapfig}
\usepackage{placeins}
\usepackage{graphicx}
\usepackage{caption}
\usepackage{mathrsfs}
\usepackage{array}
\usepackage{type1cm}
\usepackage{pifont}
\usepackage{algorithm}
\usepackage{algorithmic}

\definecolor{placeholdercolor}{RGB}{150,52,45}

\usepackage{hyperref}
\hypersetup{
  colorlinks=true,        
  citecolor=gray,         
  linkcolor=orange,       
  urlcolor=magenta        
}
\usepackage{url}

\RequirePackage{amsmath}
\RequirePackage{booktabs}
\RequirePackage{array}
\RequirePackage{siunitx}
\RequirePackage{multirow}
\RequirePackage{xcolor}
\definecolor{dyqMainAFillSixteen}{HTML}{FAF8F3}
\definecolor{dyqMainAFillEight}{HTML}{F4F9F6}
\definecolor{dyqMainAFillFour}{HTML}{F9F5FA}
\definecolor{dyqMainDynamicFill}{HTML}{F5F6F7}
\definecolor{dyqMainBlue}{HTML}{F0F5FA}
\definecolor{dyqMainGroupInk}{HTML}{555B63}
\newcommand{\dyqMainTableSetup}{%
  \fontsize{8}{10}\selectfont 
  \setlength{\tabcolsep}{0pt}%
  \renewcommand{\arraystretch}{1.00}%
  \setlength{\aboverulesep}{1.5pt}%
  \setlength{\belowrulesep}{1.5pt}%
  \setlength{\cmidrulesep}{1pt}%
  \setlength{\heavyrulewidth}{0.65pt}%
  \setlength{\lightrulewidth}{0.35pt}%
  \setlength{\cmidrulewidth}{0.25pt}%
  \sisetup{detect-weight=true,mode=text,group-digits=false,table-number-alignment=center}}
\newcommand{\dyqMainModelHead}[1]{{\small\bfseries #1}}
\newlength{\dyqMainMetricGap}
\newlength{\dyqMainModelGap}
\newcommand{\dyqMainModelSep}{\hspace{\dyqMainModelGap}}
\newcommand{\dyqMainMetricSep}{\hspace{\dyqMainMetricGap}}
\newcommand{\dyqMainMethodSep}{\hspace{1mm}}
\newcommand{\dyqMainPrecisionGroup}[2]{%
  \noalign{\vskip1.3pt\hbox{\begingroup\setlength{\fboxsep}{0pt}%
    \colorbox{#1}{\parbox[c][7.5pt][c]{\linewidth}{%
      \hspace{0.8mm}\fontsize{7.5}{7.5}\selectfont
      \color{dyqMainGroupInk}\itshape #2}}%
    \endgroup}\vskip0.5pt}}
\makeatletter
\newcommand{\dyqMainPrecisionBand}[1]{\noalign{\begingroup\color{#1}\hrule height\dimexpr\ht\@arstrutbox+\dp\@arstrutbox\relax\endgroup\vskip-\dimexpr\ht\@arstrutbox+\dp\@arstrutbox\relax}}
\makeatother

\newsavebox{\dyqMainTableBox}
\newcommand{\dyqMainRenderTable}{\usebox{\dyqMainTableBox}}
\definecolor{dyqComparisonTint}{HTML}{F0F5FA}

\newcommand{\method}{\textit{DyQ-VLA}}

\title{\textit{DyQ-VLA}: Dynamic Quantization and Compensation for VLA models with Runtime Error Awareness}

\author{%
\parbox[t]{\dimexpr\textwidth-2\tabcolsep\relax}{%
\raggedright
{\bfseries
Zihao~Zheng$^{1}$\thanks{Equal contribution.\quad
$^{\dagger}$Corresponding author: \texttt{xiang.chen@pku.edu.cn}.}\quad
Hangyu~Cao$^{2}$\footnotemark[1]\quad
Chibang~Tao$^{3}$\quad
Zhihao~Mao$^{1}$\\
Lingyue~Zhang$^{4}$\quad
Maoliang~Li$^{1}$\quad
Jiayu~Chen$^{1}$\quad
Hailong~Zou$^{1}$\\
Xuanzhe~Liu$^{1}$\quad
Donggang~Cao$^{1}$\quad
Hong~Mei$^{1}$\quad
Xiang~Chen$^{1,\dagger}$}\\[0.15em]
{\normalfont\small
$^{1}$School of Computer Science, Peking University, Beijing, China\\
$^{2}$School of Software and Microelectronics, Peking University, Beijing, China\\
$^{3}$School of Computer Science, Northwestern Polytechnical University, Xi'an, China\\
$^{4}$School of Electronics Engineering and Computer Science, Peking University, Beijing, China}
}}

\iclrfinalcopy

\begin{document}
\maketitle
\lhead{Preprint. Under review.}

\begin{abstract}
Vision-language-action (VLA) models achieve favorable task performance, yet runtime errors in closed-loop execution evolve with alternating updates of actions and observations. 
Existing VLA quantization methods mainly trade off inference speed and model performance while rarely investigating how quantization alters runtime errors. 
By comparing error distributions between full-precision and quantized models, we identify distinct evolutionary patterns for the two types of errors: quantization enlarges the variance of translation error distributions and increases their dispersion, whereas the distribution center of rotation errors gradually shifts across execution steps, demonstrating cumulative drift. 
Motivated by this observation, we rethink the optimal quantization strategy for VLA models and propose \method{}, a runtime-error-aware quantization framework, which dynamically selects activation precision according to execution steps and tracks as well as compensates rotation errors via accumulated quantization residuals. 
Corresponding operators and runtime adaptation strategies are devised within the framework to enable dynamic-precision execution.
Experiments show that \method{} achieves a 1.88$\times$ to 1.93$\times$ inference speedups while maintaining comparable or higher average task success rates.
Moreover, it reduces mean execution steps by 17.7\% to 19.8\%.
Our code is here: \textcolor{magenta}{\url{https://anonymous.4open.science/r/DyQ-VLA-7F51/}}.
\end{abstract}
\section{Introduction}
\label{sec:introduction}




Vision-Language-Action (VLA) models unify visual perception, language understanding, and action generation within a single framework~\citep{rt_2,openvla}.
Their strong performance across diverse robotic manipulation tasks has established them as a mainstream approach to embodied intelligence~\citep{pi_0.5,survey1,survey2}.

\begin{figure}[t]
    \centering
    \includegraphics[width=5.5in]{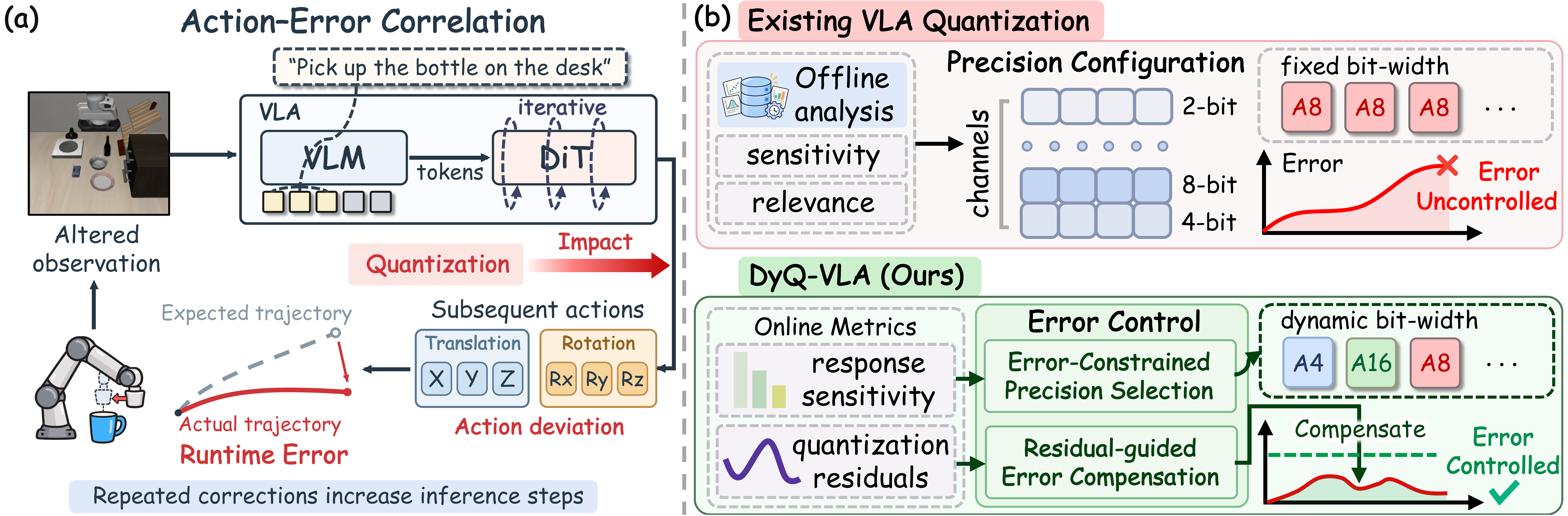}
    \caption{
    \textbf{Action-Error Correlation and \method{} Overview.}
    (a)~Runtime errors and actions interaction of VLA Models.
    (b)Runtime-Error-Aware VLA Quantization Framework -- \method{}.
    }
    \label{fig:overview}
\end{figure}

In practical VLA execution, deviations between actual and expected trajectories (e.g., human-centric demonstration trajectories) give rise to \textbf{runtime errors}, as shown in \textcolor{orange}{Fig.~\ref{fig:overview}~(a)}. 
A model may still complete the task by correcting these errors, but repeated corrections increase the number of inference steps and task completion time. 
As current errors affect subsequent observations and actions, they evolve into a dynamic \textbf{action-error correlation}. 
Existing methods, such as FLARE~\citep{flare} and REPO-VLA~\citep{repo}, seek to correct execution errors online, yet this correlation remains difficult to characterize explicitly.

Meanwhile, quantization reduces VLA memory use and inference latency, becoming a popular optimization methods for VLA real-time application~\citep{qvla,quantvla}. 
Existing approaches (e.g. QVLA~\citep{qvla}, QuantVLA~\citep{quantvla} and ActQuant~\citep{actquant}) typically assess whether faster inference preserves task success, but give less attention to how quantization changes the action-error correlation and the corrections it requires. 
We therefore ask: \textit{\textbf{How does quantization alter action-error correlation, and how can this understanding guide a quantization strategy that reduces overall task completion time?}}

In this paper, we first analyze how runtime errors evolve and how quantization changes their relationship with subsequent actions. 
Decomposing action errors into translation and rotation reveals distinct patterns: translational errors vary substantially around a relatively stable mean, while rotational errors persist more consistently across steps.
Quantization further disperses the translational error distribution but only slightly alters the rotational error distribution.
To characterize these effects at runtime, we introduce metrics for the action response to existing errors and for the accumulated rotational error. 
These findings motivate two key insights: dynamic activation precision to control translational error dispersion and residual compensation to mitigate rotational drift.

Based on the foregoing analysis, we propose \method{}, a runtime-error-aware dynamic quantization framework for VLA models. 
This framework incorporates two core mechanisms: (1) dynamic selection of activated quantization precision at each inference step with static weight precision preserved to effectively regulate translation error distribution; (2) continuous residual-based compensation for rotation error distribution during execution for effective rotation error control. 
Moreover, we implement runtime adaptations for \method{}, including dynamically precise quantization operators, efficient operator fusion strategies, and system-level scheduling optimization.

In summary, our contributions are below:
\begin{itemize}[leftmargin=*]
\item We explicitly analyze action-error correlation, identify error distribution characteristics of its different components, and reveal quantization’s effect on error distributions, offering key insights to rethink VLA quantization strategies.
\item Based on these observations, we rethink the optimal quantization strategy for VLA models and propose \method{}, the first runtime-error-aware VLA quantization framework to our knowledge. \method{} integrates a dynamic activation quantization strategy to mitigate translational error components and a residual-based compensation mechanism to suppress rotational components, with runtime optimization and adaptation.
\item We perform simulation and real-world tests on \method{} and show that it outperforms SOTA methods in runtime error control and inference speed, which further validates that VLA quantization should be investigated from the perspective of runtime error.
\end{itemize}

On popular simulation benchmarks and real-world scenarios, \method{} achieves up to 1.93$\times$ speedups and reduces mean translation and rotation runtime errors to 78.5\% - 87.0\% and 59.4\% - 82.5\%, respectively.
Moreover, it reduces mean execution steps by 17.7\% to 19.8\%.
\section{Background}
\label{tex:background}

\subsection{Runtime Error in VLA Models}

\textbf{Runtime Error Between Execution and Expected in VLA Inference.}
During VLA execution, runtime error denotes the deviation of actual trajectories from desired ones such as human demonstration trajectories. 
Due to model robustness and environmental fault tolerance, such errors do not necessarily lead to task failure, yet extra planning for error correction increases inference steps and task completion time \citep{liberox,policytrim}. 
Accordingly, VLA acceleration demands both lower per-inference cost \citep{vlacache} and fewer error-triggered extra inferences \citep{policytrim,posa}, constituting two complementary optimization directions.

\textbf{Action-Error Correlation in VLA Models.}
In VLA models, runtime errors interfere with subsequent action generation covering translation ($X$, $Y$, $Z$), rotation ($R_X$, $R_Y$, $R_Z$) and gripper control, while action execution alters future errors and forms a dynamically coupled action-error correlation.
Current research targets runtime error analysis and mitigation via learning-based methods such as FLARE~\citep{flare} and REPO-VLA~\citep{repo}, which lack explicit characterisation of action-error feedback relationships and remain under investigation regarding their stepwise evolutionary mechanisms.

\subsection{VLA Model Quantization}

\textbf{Precision Selection and Accuracy-Speed Trade-off for VLA Model Quantization.}
Quantization cuts storage and computational costs and speeds up inference by decreasing the numerical precision of model parameters, acting as an effective optimization strategy for VLA models. 
Existing research on VLA model quantization mainly explores precision selection and the accuracy-speed tradeoff, such as QVLA~\citep{qvla}, DA-PTQ~\citep{daptq}, and ActQuant~\citep{actquant}.
These methods achieve certain effects for VLA real-time deployment.

\textbf{Quantization Impact on Action-Error Correlation in VLA Models.}
However, quantization performs low-precision transformations on parameters, alters the distribution of internal representations, distorts the model’s input-output mapping, and affects runtime errors~\citep{quantvla}.
Thus, quantization raises the complexity of the relationship between errors and actions~\citep{holoqvla}, which prevents the direct application of original correlation-based error analysis and compensation to quantized models.
Current works cannot analyze quantization’s effects on runtime errors and overlook extended inference steps induced by runtime errors, leading to suboptimal speedup~\citep{volt,object}.
Additional related work is provided in Appendix~\ref{app:extended_related_work}.
\section{Runtime Error Analysis for VLA Model Quantization}
\label{sec:analysis}

\subsection{Analysis of Action-Error Correlation in Full-Precision VLA}
\label{sec:analysis_1}

\begin{figure}[t]
    \centering
    \includegraphics[width=5.5in]{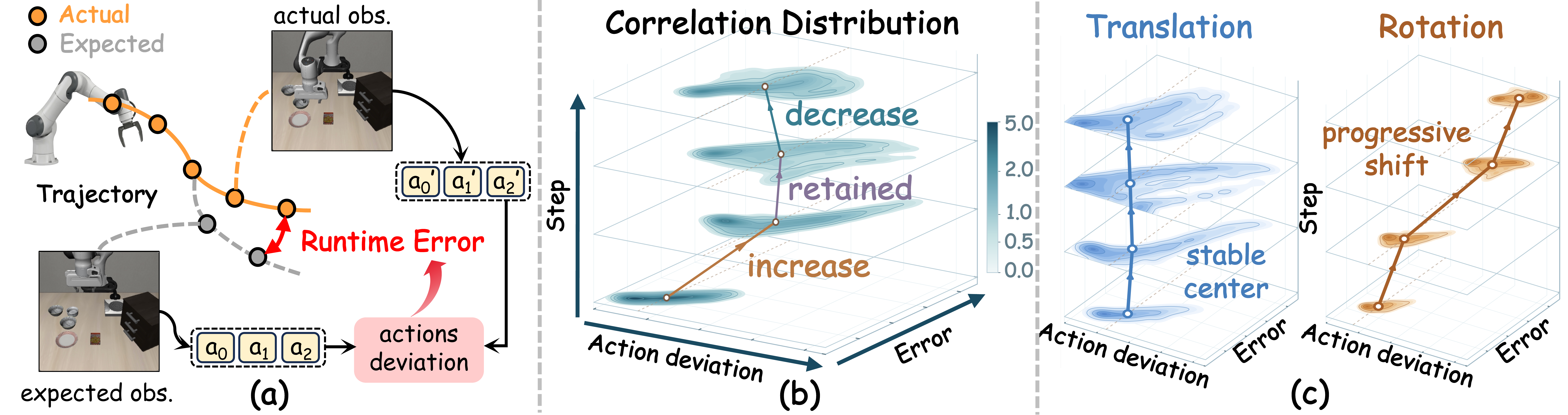}
    \caption{
    \textbf{Action–Error Correlation in Closed-Loop VLA Execution.}
    (a)~Runtime errors alter observations and subsequent actions, which in turn change runtime errors.
    (b)~Joint distributions of correlation changes in both center and shape across steps.
    (c)~The distribution for translation remains dispersed around a stable center, while rotation stays concentrated and shifts toward larger errors.
    }
    \label{fig:fig_analysis_1}
\end{figure}

\textbf{Complexity of Action-Error Correlation.}
During closed-loop VLA execution, runtime errors alter subsequent input observations and thereby affect subsequent action predictions.
Executing these actions in turn changes runtime errors, forming a complex coupled action–error correlation.
As shown in \textcolor{orange}{Fig.~\ref{fig:fig_analysis_1}~(a)}, the error at step $t$ alters the observation and predicted action sequence, resulting in a larger error at $t+H$.
The current runtime error thus depends on both the current action and historical errors. Such cross-step dependence complicates the analysis and modeling of runtime errors.

\textbf{Joint Distribution for Modeling Action-Error Correlation.}
To model action-error correlation, we investigate the joint distribution of action deviations and runtime errors over steps via repeated rollouts with equal-magnitude perturbations at a fixed execution step. 
As illustrated in \textcolor{orange}{Fig.~\ref{fig:fig_analysis_1}~(b)}, the centroid of the correlation distribution varies across steps and presents error amplification, attenuation or relative retention. 
Its width also expands or contracts with steps, and variations in both centroid and width characterize the complexity of this correlation distribution.

\textbf{Error Decomposition in terms of Translation/Rotation Distribution.}
Due to complex variations in the above distribution, further analysis is challenging. 
To reveal its inherent rules, we decompose actions into translational and rotational components and separately analyse their action-error correlation distributions. 
As illustrated in \textcolor{orange}{Fig.~\ref{fig:fig_analysis_1}~(c)}, both correlation distributions concentrate on small values initially; the rotational distribution is relatively compact, while the translational one is more dispersed. 
As execution continues, the rotational distribution stays concentrated, and its mean gradually shifts toward larger errors in a consistent direction. 
By contrast, the translational distribution maintains dispersion around a stable mean throughout all steps. 
These observations demonstrate divergent evolution patterns of action-error correlation for different motion components.

\textit{\textbf{Observation \ding{172}:}
Action-Error Correlation exhibits distinct component-wise patterns: the translational distribution has high variance and a relatively stable central value, while the rotational distribution has low variance and a center that progressively shifts in the same direction toward larger errors.}

\subsection{Quantization Impact on Action-Error Correlation in VLA Models}
\label{sec:analysis_2}

\textbf{Sensitivity of Translation/Rotation Distribution to Quantization.}
However, quantization will further affects aforementioned correlation.
To investigate the impact of quantization on such correlation, we group samples by action direction and compare error distributions before and after quantization at each step.
We define sensitivity as the change in subsequent action predictions per unit runtime error as shown in \textcolor{orange}{Eq.~(\ref{eq:error_sensitivity})}.
Here, $\boldsymbol{A}^{(j)}$ and $\overline{\boldsymbol{A}}$ denote predictions from matched perturbed and unperturbed executions at the same precision, $h$ is the prediction horizon, and $\varepsilon_t^{(j)}$ is the normalized error magnitude in direction $j$.
Differences in sensitivity before and after quantization measure its impact on action-error correlation.
\begin{equation}
\mathscr{S}_{t,j} = \frac{1}{\varepsilon_t^{(j)}}
\sqrt{
\frac{1}{h}
\sum_{i=1}^{h}
\left\|
\boldsymbol{A}_{t+1,i}^{(j)}
-
\overline{\boldsymbol{A}}_{t+1,i}
\right\|_2^2
}.
\label{eq:error_sensitivity}
\end{equation}

\begin{figure}[t]
    \centering
    \includegraphics[width=5.5in]{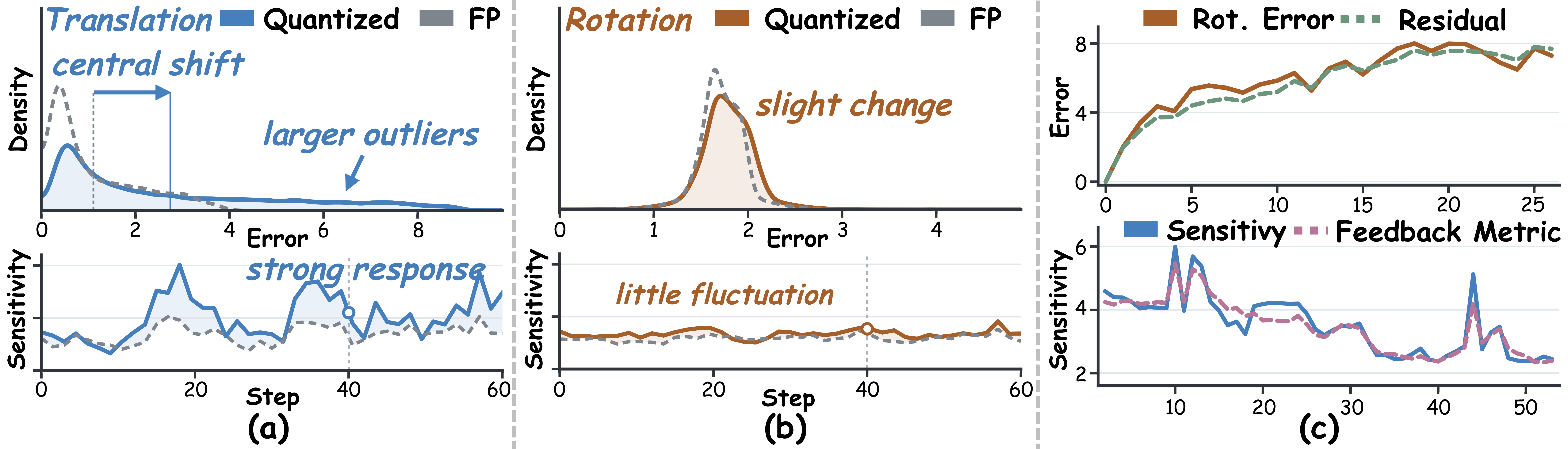}
    \caption{\textbf{Quantization Effects and Online Tracking.}
    (a)~Quantization increases translation error variance and response sensitivity.
    (b)~Rotation error distributions and sensitivity change only slightly.
    (c)~Residual and feedback metrics track rotational error and translational sensitivity, respectively.
    }
    \label{fig:fig_analysis_2}
\end{figure}

\textbf{Quantization Impact on the Translation Error Distribution.}
\textcolor{orange}{Fig.~\ref{fig:fig_analysis_2}~(a)} shows that quantization substantially alters the translation error distribution and increases its variance.
Quantization further amplifies some large errors into more pronounced outliers and shifts the distribution center toward larger errors.
Sensitivity analysis further reveals a stronger action response per unit translation error after quantization, with sensitivity varying across steps.
This indicates that quantization significantly alters the translation action-error correlation, increasing error variance and shifting the distribution center toward larger errors, with varying effects on sensitivity across steps.

\textbf{Quantization Impact on the Rotation Error Distribution.}
\textcolor{orange}{Fig.~\ref{fig:fig_analysis_2}~(b)} shows that under identical conditions, the overall distribution pattern of rotation errors changes slightly before and after quantization, with weaker amplification of extreme errors than in translation. 
Using the same sensitivity metric, we find that the action response per unit rotation error changes only slightly after quantization, with little fluctuation across execution steps.
This indicates that quantization has a limited effect on the rotation action-error correlation, with smaller distribution changes and weaker single-step response amplification than in translation.

\textit{\textbf{Observation \ding{173}:}
Quantization increases translation error variance and response sensitivity, with its impact on sensitivity varying across steps. In contrast, rotation error distributions and response sensitivity change only slightly after quantization.}

\subsection{Residual Metric for Quantization Impact on Action-Error Correlation}
\label{sec:analysis_3}

\textbf{Quantization Residuals for Tracking Rotation Errors.}
To support compensation for progressive shifts in the rotation error distribution, we project the difference between activations and their quantized reconstructions onto rotational action components.
We accumulate these signed projections across steps in a common coordinate frame to construct a cumulative quantization residual metric:
\begin{equation}
\boldsymbol{\mathcal R}_t^R
= \sum_{k=0}^{t-1}\boldsymbol{\zeta}_k^R,
\qquad
\bigl\|\boldsymbol{\mathcal R}_t^R\bigr\|_2^2
=
\sum_{k=0}^{t-1}\bigl\|\boldsymbol{\zeta}_k^R\bigr\|_2^2
+
2\sum_{0\leq i<j<t}
\langle\boldsymbol{\zeta}_i^R,\boldsymbol{\zeta}_j^R\rangle.
\label{eq:qdq_residual}
\end{equation}
Here, $\boldsymbol{\zeta}_k^R$ denotes the signed rotational projection of the activation quantization residual at query $k$.
Positive and negative cross terms capture reinforcement and cancellation across steps, respectively.
By preserving historical direction information, this metric provides an online proxy for the current net residual of rotation errors.

\textbf{Execution Feedback for Characterizing Response Sensitivity.}
To constrain the single-step amplification of translation errors, we estimate the response sensitivity defined in \textcolor{orange}{Sec.~\ref{sec:analysis_2}} using deviations between actual and planned translation, together with changes in rotational actions.
At each step, this estimate characterizes the change in subsequent action predictions per unit translation error, providing an online signal for constraining extreme action responses.

\textbf{Online Metrics for Guiding Quantization Control.}
\textcolor{orange}{Fig.~\ref{fig:fig_analysis_2}~(c)} presents the tracking results for both metrics.
The cumulative quantization residual closely follows rotation runtime error measured against the matched reference trajectory, while the feedback metric tracks changes in measured translation response sensitivity across steps.
Accordingly, cumulative residuals can guide compensation direction and magnitude to suppress continuous drift in the rotation error distribution.
The estimated translation response sensitivity guides per-step deviation bounds and precision selection within these bounds, helping limit further amplification of extreme translation errors.
We extend the above analyses across multiple scenarios and observe consistent patterns, supporting the generality (\textcolor{orange}{Appendix.~\ref{app:runtime_error_analysis}}).

\textit{\textbf{Key Insights for VLA Quantization Design:}
The proposed metrics characterize and track quantization effects on both distributions. We should use these metrics to guide runtime dynamic precision adjustment for translation error control and effective rotation error compensation.}
\section{Runtime Error Aware Dynamic Quantization Framework}
\label{sec:method}

\subsection{Dynamic Quantization Strategy for Translation Error Control}
\label{sec:method_1}

\textbf{Offline Error and Latency Profiling.}
Aformentioned analysis reveals that translation error distribution at each step can be controlled by regulating quantization precision, while rotation error distribution can be corrected via continuous compensation. 
First, we explore that precision selection requires error and latency estimates for candidate configurations.
Since quantization loss mainly arises from activations~\citep{awq}, we retain static weight quantization and profile activation precision configurations offline, avoiding per-candidate evaluation at runtime.
Sepcifically, we estimate $E_d(b)$ using high quantiles of translational and rotational errors and record inference latency $L(b)$.
The reference configuration provides the calibration error baseline $\tau_d$.

\textbf{Runtime-Error-Aware Quantization Precision Selection.}
Our analysis shows that quantization increases translation error variance, making the distribution more dispersed and harder to track.
Existing studies overlook this issue and use static activation quantization.
We therefore dynamically select activation precision to constrain this variance at each step.
Before inference $t$, we estimate the normalized response sensitivity $\widehat{s}_{t,d}$ from the previous execution's translation shortfall relative to its plan and rotation variations in the existing motion plan.
Using offline profiles, we select the configuration minimizing inference latency under translation and rotation error constraints:
\begin{equation}
\underset{b\in\mathcal{B}}{\arg\min}\; L(b)
\quad\text{s.t.}\quad
\max_{d\in\{T,R\}}
\frac{
\widehat{s}_{t,d}\bigl(E_d(b)+r_{t-1,d}\bigr)
}{
\tau_d
}
\leq 1.
\label{eq:4-1}
\end{equation}
Here, $\mathcal{B}$ is the set of candidate activation precision configurations, and $T$ and $R$ denote translation and rotation, respectively.
To account for persistent historical errors, $r_{t-1,R}$ denotes the estimated magnitude of the projected rotation residual remaining after the previous compensation, on the same scale as $E_R(b)$; $r_{t-1,T}=0$.
Higher sensitivity or larger residuals tighten the error bound.
Selection uses table lookup and constraint checks, with a predefined fallback if no candidate is feasible.
Together, these mechanisms mitigate quantization effects on the translation error distribution while reducing unnecessary high-precision computation.

\begin{figure}[t]
    \centering
    \includegraphics[width=5.5in]{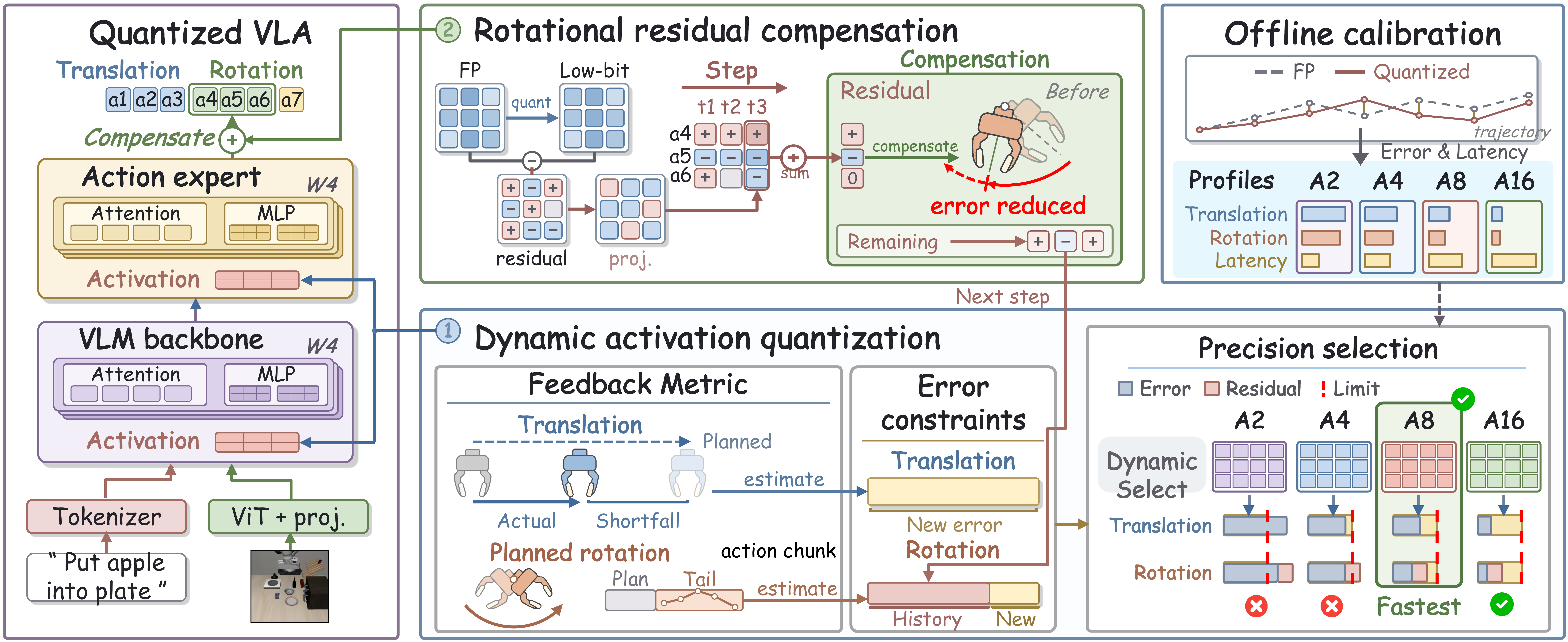}
    \caption{
    \textbf{Framework of \method{}.}
    Runtime feedback guides dynamic activation quantization for translation error control, while cumulative residuals guide rotation error compensation.
    }
    \label{fig:framework}
\end{figure}

\subsection{Residual Compensation Strategy for Rotation Error Control}
\label{sec:method_2}

\textbf{Historical Rotation Residual Estimation.}
Our analysis shows that the rotation error distribution is concentrated and little affected by quantization, yet its center continuously drifts in one direction, causing large accumulated errors late in execution.
Existing methods apply global compensation, overlooking differences in distribution evolution across motion components and risking under- or over-compensation.
To determine the correction direction and magnitude, we estimate the current net residual of historical rotation errors using the cumulative quantization residual metric in \textcolor{orange}{Sec.~\ref{sec:analysis_3}}.

\textbf{Residual-Guided Action Compensation.}
Using this signed residual estimate, we apply corrections in the opposite direction to counteract the drift.
We express the rotational residual estimate $\widehat{\mathbf{e}}_t$ and rotation command $\mathbf{a}_{t,i}$ in a common coordinate system and normalized scale.
To respect action bounds, we limit the compensation ratio and distribute the correction across the $h$ actions executed in the current round, as shown in \textcolor{orange}{Eq.~(\ref{eq:4-2})}. 
Here, $\eta\in(0,1]$ controls compensation strength, and absolute values and inequalities are evaluated componentwise.
To avoid repeated compensation, we update the historical residual estimate using the actual rotation command corrections.
The remaining estimated residual guides the next round of precision selection and action compensation.
Together, residual estimation and bounded action compensation suppress persistent drift of the rotation error distribution, keeping its center at a low error level.
\begin{equation}
\alpha_t
=
\eta \max_{0 \leq \alpha \leq 1}
\left\{
\alpha :
|\mathbf{a}_{t,i}| + \frac{\alpha}{h}|\widehat{\mathbf{e}}_t|
\leq \mathbf{1},\ \forall i
\right\},
\qquad
\widetilde{\mathbf{a}}_{t,i}
=
\mathbf{a}_{t,i}
-
\frac{\alpha_t}{h}\widehat{\mathbf{e}}_t,
\quad i=1,\ldots,h.
\label{eq:4-2}
\end{equation}

\subsection{Runtime Optimization for \textit{DyQ-VLA} Framework}
\label{sec:method_3}

\textbf{Dynamic Quantization Kernels with Shared Weights.}
Dynamic activation precision with fixed W4 weights requires matching different Tensor Core operand formats while preserving shared weight storage.
We implement dedicated A2/A4/A8/A16 kernels: A2 uses structured sparse INT4 computation, while A4 uses dense INT4 computation.
The A8 path expands W4 fragments to INT8 in registers, and A16 reconstructs W4 tiles as BF16 in shared memory.
All paths access the same resident W4 allocation through zero-copy views, keeping format conversions on chip and avoiding additional global weight copies or repacking.

\textbf{Kernel Flow Fusion and System-Level Scheduling Optimization.}
To reduce intermediate data movement, we fuse dequantization and compatible RoPE, GELU, gating, and residual operations with GEMM, using epilogue fusion where applicable.
Intermediate results stay in registers or shared memory where possible, reducing HBM writes and reads.
Fixed-shape kernels and preallocated buffers support CUDA Graph replay of the selected precision path, reducing repeated scheduling and host kernel launch overhead.
Details and illustrations of overall runtime optimization are presented in \textcolor{orange}{Appendix~\ref{app:method_details}}.
Together, these optimizations enable low-overhead precision switching and translate dynamic quantization into practical GPU acceleration.

\section{Experiments}
\label{tex:experiments}

\subsection{Experimental Setup}
We evaluate \method{} on $\pi_{0.5}$~\citep{pi_0.5} and GR00T-N1.5~\citep{gr00t} on LIBERO~\citep{libero}, with additional $\pi_{0.5}$ tests on RoboTwin 2.0~\citep{robotwin} and an Agilex Piper-X arm.
We compare against BF16, five general PTQ methods~\citep{gptq,awq,omniquant,smoothquant,duquant}, and four VLA quantization baselines~\citep{actquant,qvla,quantvla,holoqvla} (\textcolor{orange}{Tab.~\ref{tab:main_results}}).
All experiments run on an NVIDIA A100 GPU; \method{} uses W4 weights with dynamic A2/A4/A8/A16 activations.
SR covers all rollouts; execution steps and end-effector errors relative to matched demonstrations are averaged over successful episodes.
See more details in \textcolor{orange}{Appendix~\ref{app:error_measurement}}.

\subsection{Evaluating Results}
\begin{table}[t]
\caption{\textbf{Quantization Results on the LIBERO Simulation Benchmark.}}
\label{tab:main_results}
\centering
\begingroup

\dyqMainTableSetup
\addtolength{\dyqMainModelGap}{-1.2mm}
\vspace{2pt}

\begin{lrbox}{\dyqMainTableBox}%
\begin{tabular}{
@{}
>{\raggedright\arraybackslash}p{16.1mm}
@{\dyqMainMethodSep}
>{\centering\arraybackslash}p{6.4mm}
>{\centering\arraybackslash}p{7.8mm}
>{\centering\arraybackslash}p{6.4mm}
>{\centering\arraybackslash}p{6.4mm}
>{\centering\arraybackslash}p{6.4mm}
@{\dyqMainMetricSep}
>{\centering\arraybackslash}p{7.8mm}
@{\dyqMainMetricSep}
>{\centering\arraybackslash}p{7.8mm}
@{\hspace{0.6mm}}
>{\centering\arraybackslash}p{7.8mm}
@{\dyqMainModelSep}
>{\centering\arraybackslash}p{6.4mm}
>{\centering\arraybackslash}p{7.8mm}
>{\centering\arraybackslash}p{6.4mm}
>{\centering\arraybackslash}p{6.4mm}
>{\centering\arraybackslash}p{6.4mm}
@{\dyqMainMetricSep}
>{\centering\arraybackslash}p{7.8mm}
@{\dyqMainMetricSep}
>{\centering\arraybackslash}p{7.8mm}
@{\hspace{0.6mm}}
>{\centering\arraybackslash}p{7.8mm}
@{}
}
\toprule
\multirow{3}{16.1mm}[-3.5pt]{\centering\bfseries Method}
& \multicolumn{8}{@{}c@{\dyqMainModelSep}}
  {\dyqMainModelHead{$\boldsymbol{\pi}_{0.5}$}}
& \multicolumn{8}{@{}c@{}}
  {\dyqMainModelHead{GR00T-N1.5}}
\\
\cmidrule(r{\dyqMainModelGap}){2-9}
\cmidrule{10-17}
&
\multicolumn{5}{@{}c@{\dyqMainMetricSep}}
  {Success rate (\%) $\uparrow$}
&
\multicolumn{1}{@{}c@{\dyqMainMetricSep}}
  {\multirow{2}{7.8mm}[-1.95pt]{\centering Steps$\downarrow$}}
&
\multicolumn{2}{@{}c@{\dyqMainModelSep}}
  {\bfseries Error $\downarrow$}
&
\multicolumn{5}{@{}c@{\dyqMainMetricSep}}
  {Success rate (\%) $\uparrow$}
&
\multicolumn{1}{@{}c@{\dyqMainMetricSep}}
  {\multirow{2}{7.8mm}[-1.95pt]{\centering Steps$\downarrow$}}
&
\multicolumn{2}{@{}c@{}}
  {\bfseries Error $\downarrow$}
\\
\cmidrule(r{\dyqMainMetricGap}){2-6}
\cmidrule(r{\dyqMainModelGap}){8-9}
\cmidrule(r{\dyqMainMetricGap}){10-14}
\cmidrule{16-17}
& {Spa.} & {Obj.} & {Goal} & {Long} & {\bfseries Avg.}
& {} & {Trans.} & {Rot.}
& {Spa.} & {Obj.} & {Goal} & {Long} & {\bfseries Avg.}
& {} & {Trans.} & {Rot.}
\\
\midrule
BF16 ref.
& 98.4 & 97.8 & 98.4 & 93.4 & 97.0
& 159.7
& 60.37 & 14.19
& 87.4 & 91.8 & 89.4 & 75.8 & 86.1
& 237.2
& 85.40 & 18.35
\\
\dyqMainPrecisionGroup{dyqMainAFillSixteen}{W4A16}
GPTQ
& 15.0 & 39.0 & 10.0 & 0.0 & 16.0
& 248.0
& 161.19 & 27.62
& 43.0 & 73.0 & 49.0 & 60.0 & 56.3
& 320.0
& 160.32 & 26.91
\\
AWQ
& 4.0 & 29.0 & 11.0 & 2.0 & 11.5
& 255.0
& 171.22 & 29.26
& 38.0 & 74.0 & 51.0 & 64.0 & 56.8
& 322.0
& 167.18 & 28.03
\\
OmniQuant
& 8.0 & 23.0 & 10.0 & 0.0 & 10.3
& 260.0
& 176.23 & 29.93
& 39.0 & 65.0 & 43.0 & 53.0 & 50.0
& 338.0
& 176.80 & 29.52
\\
ActQuant
& \bfseries 98.8
& {\underline{99.0}}
& {\underline{96.4}}
& 91.6
& 96.5
& 191.2
& 72.08
& 15.52
& 87.8
& 93.0
& 86.8
& 73.8
& 85.4
& 275.1
& 102.56
& {\underline{19.65}}
\\
QVLA
& 98.0
& 97.2
& {\underline{96.4}}
& 91.8
& 95.9
& 193.0
& 72.58
& {\underline{14.56}}
& 86.0
& 90.4
& 86.6
& 72.6
& 83.9
& 272.5
& 106.46
& 20.35
\\
\dyqMainPrecisionGroup{dyqMainAFillEight}{W4A8}
SmoothQuant
& 98.2
& 98.4
& 95.6
& 93.4
& 96.4
& 169.6
& 72.50
& 18.81
& 82.6
& 91.6
& 88.2
& 78.8
& 85.3
& {\underline{255.4}}
& 103.16
& 22.31
\\
DuQuant
& 94.6
& 98.0
& 93.8
& 91.0
& 94.4
& {\underline{168.2}}
& 74.94
& 19.06
& 73.4
& 86.6
& 81.2
& 77.8
& 79.8
& 263.0
& 113.46
& 23.05
\\
QuantVLA
& {\underline{98.7}}
& 98.0
& \bfseries 98.0
& \bfseries 96.0
& {\underline{97.7}}
& 186.4
& {\underline{70.33}}
& 18.17
& \bfseries 93.6
& {\underline{93.2}}
& \bfseries 90.4
& 73.4
& {\underline{87.7}}
& 267.2
& {\underline{97.77}}
& 21.86
\\
\dyqMainPrecisionGroup{dyqMainAFillFour}{W4A4}
SmoothQuant
& 83.0
& 88.0
& 40.0
& 26.0
& 59.3
& 211.0
& 139.02
& 28.22
& 85.2
& 84.8
& 81.6
& {\underline{79.4}}
& 82.8
& 259.0
& 147.25
& 28.80
\\
DuQuant
& 96.2
& 98.2
& 93.4
& 88.6
& 94.1
& 171.5
& 91.47
& 20.75
& 64.4
& 74.8
& 63.2
& 79.0
& 70.4
& 284.0
& 148.33
& 28.89
\\
QuantVLA
& 96.0
& \bfseries 100.0
& \bfseries 98.0
& 86.0
& 95.0
& 168.6
& 85.19
& 19.88
& 63.2
& 71.8
& 60.8
& 67.4
& 65.8
& 269.5
& 154.74
& 29.81
\\
HoloQ-VLA
& 97.0
& \bfseries 100.0
& 96.0
& {\underline{94.0}}
& 96.8
& 189.6
& 84.72
& 16.64
& 86.8
& 90.8
& {\underline{90.0}}
& 79.2
& 86.7
& 263.8
& 106.84
& 20.40
\\
\dyqMainPrecisionGroup{dyqMainDynamicFill}
{W4 / Dynamic activation precision}
\dyqMainPrecisionBand{dyqMainBlue}
\textbf{\textit{DyQ-VLA}}
& \bfseries 98.8
& 98.4
& \bfseries 98.0
& \bfseries 96.0
& \bfseries 97.8
& \bfseries 149.5
& \bfseries 55.20
& \bfseries 13.23
& {\underline{89.4}}
& \bfseries 93.4
& \bfseries 90.4
& \bfseries 80.8
& \bfseries 88.5
& \bfseries 219.8
& \bfseries 85.07
& \bfseries 18.04
\\
\bottomrule
\multicolumn{17}{l}{\footnotesize $^{*}$Note: Translation/rotation errors are in mm/degrees. \textbf{Bold}/\underline{underline} denote the best/second-best results.}
\end{tabular}%
\end{lrbox}%

\dyqMainRenderTable

\endgroup
\end{table}

\textbf{Simulation Results.}
On LIBERO (\textcolor{orange}{Tab.~\ref{tab:main_results}}), \method{} achieves the highest average SR among quantized methods: 97.8\% on $\pi_{0.5}$ and 88.5\% on GR00T-N1.5, exceeding the corresponding BF16 results by 0.8 and 2.4 percentage points, respectively.
Compared with QuantVLA W4A8, the strongest baseline by average SR (97.7\% and 87.7\%, respectively), \method{} maintains comparable performance on $\pi_{0.5}$, while its improvement on GR00T-N1.5 comes primarily from Long, where SR increases from 73.4\% to 80.8\%.
On RoboTwin 2.0 (\textcolor{orange}{Tab.~\ref{tab:vla_comparison_speedup_spaced}}), $\pi_{0.5}$ with \method{} achieves 82.9\% and 76.5\% SR under Clean and Randomized settings, respectively.
These results surpass the strongest quantized baseline in each setting---QuantVLA W4A8 and HoloQ-VLA---by 1.5 and 1.3 percentage points, respectively.
Relative to BF16, SR is 0.2 percentage points higher under Clean and 0.3 percentage points lower under Randomized. 
These results support the ability of \method{} to maintain task performance across different backbones and simulation environments.

\method{} reduces rollout inference cost through fewer execution
steps.
\textcolor{orange}{Tab.~\ref{tab:vla_comparison_speedup_spaced}} compares complete inference systems, including quantization methods and operator implementations, with rollout speedup measured from cumulative model inference time.
Lower per-step latency supports timely policy updates, but quantization-induced deviations may require additional corrective steps even when policy robustness and task tolerance preserve success.
For $\pi_{0.5}$ on LIBERO, ActQuant and QuantVLA achieve per-step speedups of 1.10$\times$ and 1.32$\times$, yet longer executions reduce their rollout speedups to 0.92$\times$ and 1.13$\times$, respectively.

Compared with QuantVLA W4A8, \method{} reduces mean execution steps per successful episode by 19.8\%, 17.7\%, and 18.9\% for $\pi_{0.5}$ and GR00T-N1.5 on LIBERO and $\pi_{0.5}$ on RoboTwin 2.0, respectively.
Steps also decrease by 6.4\%, 7.3\%, and 5.3\% relative to BF16.
BF16 itself deviates from the reference demonstrations, so full precision does not guarantee the fewest execution steps.
Combining fewer steps with faster per-step inference, \method{} achieves measured rollout speedups of 1.88$\times$-1.93$\times$ over BF16, exceeding its per-step speedups of 1.74$\times$-1.81$\times$.
Rollout inference is 70.8\%-72.5\% faster than QuantVLA W4A8, the fastest baseline, at comparable static full-model weight footprints on LIBERO: 3.49 versus 3.50 GB for $\pi_{0.5}$ and 4.26 versus 4.27 GB for GR00T-N1.5.

\begin{table}[!t]
\caption{\textbf{Task Performance and Inference Efficiency
on RoboTwin 2.0 and LIBERO.}}
\label{tab:vla_comparison_speedup_spaced}
\centering
\begingroup
\fontsize{8}{10.5}\selectfont
\setlength{\tabcolsep}{0pt}
\renewcommand{\arraystretch}{1.05}
\setlength{\aboverulesep}{1.2pt}
\setlength{\belowrulesep}{1.2pt}
\setlength{\cmidrulesep}{1pt}
\setlength{\heavyrulewidth}{0.65pt}
\setlength{\lightrulewidth}{0.35pt}
\setlength{\cmidrulewidth}{0.25pt}
\providecolor{dyqMainBlue}{HTML}{F0F5FA}
\begin{tabular}{
@{}
>{\raggedright\arraybackslash}p{14.4mm}
@{\hspace{0.8mm}}
>{\centering\arraybackslash}p{10.25mm}
@{\hspace{1mm}}
>{\centering\arraybackslash}p{4.95mm}
@{\hspace{0.5mm}}
>{\centering\arraybackslash}p{7.05mm}
@{\hspace{0.5mm}}
>{\centering\arraybackslash}p{6.35mm}
@{\hspace{1.2mm}}
>{\centering\arraybackslash}p{4.95mm}
@{\hspace{0.5mm}}
>{\centering\arraybackslash}p{7.05mm}
@{\hspace{0.5mm}}
>{\centering\arraybackslash}p{6.35mm}
@{\hspace{1.2mm}}
>{\centering\arraybackslash}p{7.65mm}
@{\hspace{0.6mm}}
>{\centering\arraybackslash}p{5.05mm}
@{\hspace{0.7mm}}
>{\centering\arraybackslash}p{8.5mm}
@{\hspace{1.6mm}}
>{\centering\arraybackslash}p{5.05mm}
@{\hspace{0.6mm}}
>{\centering\arraybackslash}p{8.5mm}
@{\hspace{0.7mm}}
>{\centering\arraybackslash}p{8.2mm}
@{\hspace{1.8mm}}
>{\centering\arraybackslash}p{5.05mm}
@{\hspace{0.6mm}}
>{\centering\arraybackslash}p{8.5mm}
@{\hspace{0.7mm}}
>{\centering\arraybackslash}p{8.35mm}
@{}
}
\toprule
\multirow{3}{14.4mm}[-2.5pt]{\centering\bfseries Method}
& \multirow{3}{10.25mm}[-2.5pt]{\centering\bfseries Prec.}
& \multicolumn{9}{@{}c@{\hspace{1.6mm}}}{{\fontsize{9}{11}\selectfont\bfseries RoboTwin 2.0 / $\boldsymbol{\pi}_{0.5}$}}
& \multicolumn{6}{@{}c@{}}{{\fontsize{9}{11}\selectfont\bfseries LIBERO}}\\
\cmidrule(r{1.6mm}){3-11}\cmidrule{12-17}
& & \multicolumn{3}{@{}c@{\hspace{1.2mm}}}{Clean}
& \multicolumn{3}{@{}c@{\hspace{1.2mm}}}{Randomized}
& \multirow{2}{7.65mm}[-1.5pt]{\centering Steps$\downarrow$}
& \multicolumn{2}{@{}c@{\hspace{1.6mm}}}{Speedup$\uparrow$}
& \multicolumn{3}{@{}c@{\hspace{1.8mm}}}{{\fontsize{9}{11}\selectfont$\boldsymbol{\pi}_{0.5}$}}
& \multicolumn{3}{@{}c@{}}{{\fontsize{9}{11}\selectfont\bfseries GR00T-N1.5}}\\
\cmidrule(r{1.2mm}){3-5}\cmidrule(r{1.2mm}){6-8}
\cmidrule(r{1.6mm}){10-11}\cmidrule(r{1.8mm}){12-14}\cmidrule{15-17}
& & SR$\uparrow$ & Trans. & Rot.
& SR$\uparrow$ & Trans. & Rot. &
& Step & Rollout
& Step & Rollout & Mem.$\downarrow$
& Step & Rollout & Mem.$\downarrow$\\
\midrule
BF16 ref. & BF16
& 82.7 & 32.53 & 8.77
& 76.8 & 39.82 & 12.27
& 155.7
& 1.00 & 1.00
& 1.00 & 1.00 & 6.71
& 1.00 & 1.00 & 5.45
\\
\midrule
ActQuant & W4A16
& 80.9 & 41.65 & 12.10
& 74.2 & 52.38 & 16.48
& 186.4
& 1.11 & 0.93
& 1.10 & 0.92 & \textbf{2.63}
& 1.09 & 0.94 & \underline{4.03}
\\
QVLA & W4A16
& 80.5 & 42.18 & \underline{10.92}
& 73.7 & 51.74 & \underline{15.28}
& 188.2
& 1.09 & 0.90
& 1.08 & 0.89 & \underline{2.70}
& 1.07 & 0.93 & 4.10
\\
QuantVLA & W4A8
& \underline{81.4} & \underline{40.36} & 15.24
& 74.9 & \underline{50.81} & 20.31
& \underline{181.7}
& \underline{1.30} & \underline{1.11}
& \underline{1.32} & \underline{1.13} & 3.50
& \underline{1.23} & \underline{1.09} & 4.27
\\
HoloQ-VLA & W4A4
& 81.0 & 46.73 & 13.57
& \underline{75.2} & 58.92 & 18.61
& 184.8
& 1.16 & 0.98
& 1.17 & 0.99 & 3.54
& 1.10 & 0.99 & \textbf{3.99}
\\
\midrule
\noalign{\begingroup\color{dyqMainBlue}\hrule height11.025pt\endgroup\vskip-11.025pt}
\textbf{\method{}}
& W4/A\textsubscript{dyn}
& \textbf{82.9} & \textbf{34.08} & \textbf{9.06}
& \textbf{76.5} & \textbf{42.37} & \textbf{12.68}
& \textbf{147.4}
& \textbf{1.80} & \textbf{1.90}
& \textbf{1.81} & \textbf{1.93} & 3.49
& \textbf{1.74} & \textbf{1.88} & 4.26
\\
\bottomrule
\end{tabular}%

\endgroup
\end{table}

\begin{figure}[t]
    \centering
    \includegraphics[width=5.5in]{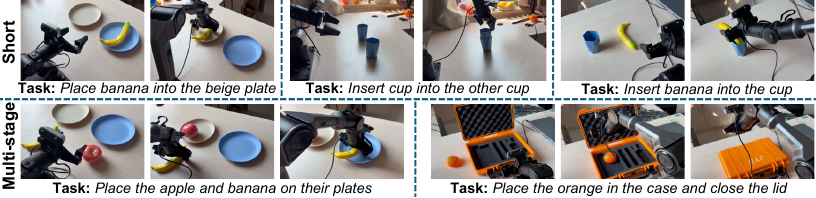}
    \caption{\textbf{Real-World Setup and Task Examples.} Details can be found in Appendix~\ref{app:real_world_details}.}
    \label{fig:real_world_setup}
\end{figure}
\FloatBarrier
\begin{table}[!htbp]
\caption{Real-world performance of \method{}, quantization baselines, and the BF16 reference.}
\label{tab:real_world}
\centering
\begingroup
\fontsize{9}{10.8}\selectfont
\setlength{\tabcolsep}{0pt}
\renewcommand{\arraystretch}{1.02}
\setlength{\aboverulesep}{1.3pt}
\setlength{\belowrulesep}{1.3pt}
\setlength{\cmidrulesep}{1pt}
\setlength{\heavyrulewidth}{0.65pt}
\setlength{\lightrulewidth}{0.35pt}
\setlength{\cmidrulewidth}{0.25pt}

\begin{tabular}{@{}
>{\raggedright\arraybackslash}p{17.4mm}
@{\hspace{1.2mm}}
>{\centering\arraybackslash}p{14.8mm}
@{\hspace{1.6mm}}
>{\centering\arraybackslash}p{11.8mm}
@{\hspace{0.4mm}}
>{\centering\arraybackslash}p{11.8mm}
@{\hspace{0.4mm}}
>{\centering\arraybackslash}p{11.8mm}
@{\hspace{0.4mm}}
>{\centering\arraybackslash}p{11.8mm}
@{\hspace{2.4mm}}
>{\centering\arraybackslash}p{14mm}
@{\hspace{0.8mm}}
>{\centering\arraybackslash}p{14mm}
@{\hspace{2.4mm}}
>{\centering\arraybackslash}p{10.6mm}
@{\hspace{1mm}}
>{\centering\arraybackslash}p{11.1mm}
@{}}

\toprule
\multirow{3}{17.4mm}[-1.2pt]{\centering\bfseries Method}
& \multirow{3}{14.8mm}[-1.2pt]{\centering Prec.}
& \multicolumn{4}{@{}c@{\hspace{2.4mm}}}{Short-horizon}
& \multicolumn{2}{@{}c@{\hspace{2.4mm}}}{Multi-stage}
& \multirow{3}{10.6mm}[-1.2pt]{\centering SR (\%)$\uparrow$}
& \multirow{3}{11.1mm}[-1.2pt]{%
    \centering\shortstack{Task\\Spd.$\uparrow$}}\\

\cmidrule(r{2.4mm}){3-6}
\cmidrule(r{2.4mm}){7-8}
& & Banana & Cup & Banana & Apple & Orange & Ordered & &\\
& & Plate & Insertion & Cup & Plate & Packing & Placement & &\\

\midrule
BF16 ref. & BF16
& 9/10 & 6/10 & 8/10 & 8/10 & 7/10 & 7/10 & 75.0 & 1.00\\

\midrule
QuantVLA & W4A8
& 8/10 & 6/10 & 8/10 & 8/10 & 6/10 & 5/10
& 68.3 & \underline{1.16}\\

ActQuant & W4A16
& 8/10 & 6/10 & 8/10 & 8/10 & 6/10 & 5/10 & 68.3 & 1.08\\

QVLA & W4A16
& 8/10 & 4/10 & 8/10 & 8/10 & 5/10 & 5/10 & 63.3 & 1.04\\

HoloQ-VLA & W4A4
& 8/10 & 6/10 & 8/10 & 8/10 & 7/10 & 6/10
& \underline{71.7} & 1.11\\

\noalign{%
  \begingroup
  \color{dyqMainBlue}
  \hrule height11.016pt
  \endgroup
  \vskip-11.016pt
}
\textbf{\textit{\method{}}}
& $\mathrm{W4}/\mathrm{A}_{\mathrm{dyn}}$
& 8/10 & 7/10 & 8/10 & 7/10 & 7/10 & 7/10
& \textbf{73.3} & \textbf{1.59}\\

\bottomrule
\end{tabular}
\endgroup
\end{table}

\textbf{Real-World Evaluation.}
Across six real-world tasks spanning short-horizon and multi-stage
manipulation (\textcolor{orange}{Tab.~\ref{tab:real_world}}), \method{} achieves the highest average SR among quantized methods at 73.3\%, close to BF16's 75.0\%.
On each of the two multi-stage tasks, \method{} succeeds in 7 out of 10 trials, matching BF16.
While maintaining this task performance, \method{} achieves a 1.59$\times$ end-to-end task completion speedup over BF16, compared with 1.16$\times$ for QuantVLA, the fastest quantized baseline, demonstrating its efficiency advantage in real-world execution.

\subsection{Discussion}

\par\noindent
\begin{minipage}{\linewidth}
\begin{minipage}[t]{\dimexpr\linewidth-54.1mm-2.5mm\relax}
\vspace{0pt}
\textbf{Runtime Error Evaluation.}
\textcolor{orange}{Tab.~\ref{tab:runtime_suite}} shows that \method{} achieves
the lowest translation and rotation errors and requires the
fewest steps on average across all four LIBERO suites, with
pronounced gains on long-horizon tasks.
On LIBERO-Long, \method{} and QuantVLA W4A8 both achieve
96.0\% SR, but \method{} reduces translation and rotation
errors relative to demonstrations by 32.1\% and 47.2\%,
respectively, and execution steps from 224.5 to 163.3,
a 27.3\% reduction.
\textcolor{orange}{Fig.~\ref{fig:efficiency}} plots SR
against the execution step budget.
Under a limited budget, \method{} achieves higher SR,
indicating that more tasks are completed within the same
number of steps.
The curve complements the final SR comparison by showing
differences in execution length, consistent with the lower
step counts reported in the table.

\end{minipage}\hspace{2.5mm}%
\begin{minipage}[t]{54.1mm}
\vspace{0pt}
\setlength{\parskip}{0pt}
\centering
\includegraphics[width=\linewidth]{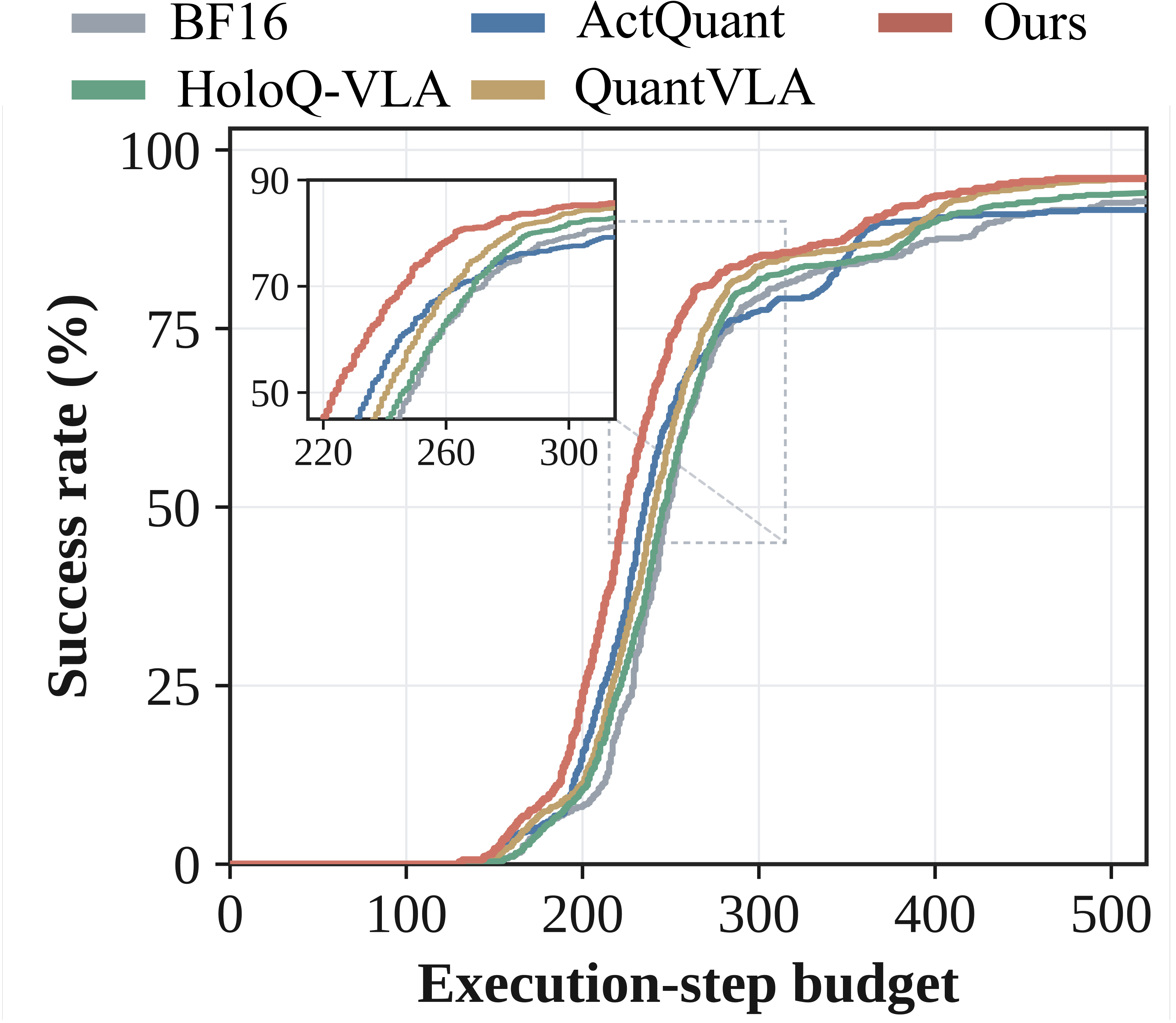}
\captionsetup{position=bottom}
\captionof{figure}{SR versus steps.}
\label{fig:efficiency}
\end{minipage}
\end{minipage}
\par

\FloatBarrier

\par\noindent\begin{minipage}{\linewidth}
\centering
\captionsetup{position=top}

\begin{minipage}[t]{83mm}
\vspace{0pt}
\captionof{table}{Runtime evaluation on LIBERO ($\pi_{0.5}$).}
\label{tab:runtime_suite}
\centering
\begingroup
\fontsize{8}{9.7}\selectfont
\renewcommand{\arraystretch}{1}
\setlength{\tabcolsep}{0pt}
\setlength{\aboverulesep}{1.1pt}
\setlength{\belowrulesep}{1.2pt}
\setlength{\cmidrulesep}{.5pt}
\setlength{\heavyrulewidth}{.6pt}
\setlength{\lightrulewidth}{.3pt}
\setlength{\cmidrulewidth}{.25pt}
\providecolor{dyqMainBlue}{HTML}{F0F5FA}
\def\runtimeHead#1#2{#1\kern.3pt\scalebox{.85}{#2}}
\makeatletter
\def\runtimeBand{\noalign{\begingroup\color{dyqMainBlue}\hbox{\vrule width\linewidth height\dimexpr\ht\@arstrutbox+\dp\@arstrutbox\relax}\endgroup\vskip-\dimexpr\ht\@arstrutbox+\dp\@arstrutbox\relax}}
\makeatother
\begin{tabular*}{\linewidth}{@{\hspace{.4mm}\extracolsep{\fill}}
>{\raggedright\arraybackslash}p{14.7mm}
@{\hspace{.7mm}}
>{\centering\arraybackslash}p{7.2mm}
@{\hspace{.4mm}}
>{\centering\arraybackslash}p{9mm}
@{\hspace{.4mm}}
>{\centering\arraybackslash}p{7mm}
@{\hspace{.4mm}}
>{\centering\arraybackslash}p{8.5mm}
@{\hspace{1mm}}
>{\centering\arraybackslash}p{7.2mm}
@{\hspace{.4mm}}
>{\centering\arraybackslash}p{9mm}
@{\hspace{.4mm}}
>{\centering\arraybackslash}p{7mm}
@{\hspace{.4mm}}
>{\centering\arraybackslash}p{8.5mm}
@{\hspace{.4mm}}}
\toprule
\multirow{2}{*}[-1pt]{\textbf{Method}} & \multicolumn{4}{c}{\textbf{Spatial}} & \multicolumn{4}{c}{\textbf{Object}} \\
\cmidrule(lr{.8mm}){2-5}\cmidrule(lr{.8mm}){6-9}
 & \runtimeHead{SR}{$\uparrow$} & \runtimeHead{Trans.}{$\downarrow$} & \runtimeHead{Rot.}{$\downarrow$} & \runtimeHead{Steps}{$\downarrow$} & \runtimeHead{SR}{$\uparrow$} & \runtimeHead{Trans.}{$\downarrow$} & \runtimeHead{Rot.}{$\downarrow$} & \runtimeHead{Steps}{$\downarrow$} \\
\midrule
ActQuant & \textbf{98.8} & 63.82 & \underline{12.42} & 177.0 & \underline{99.0} & \underline{64.95} & 14.56 & 174.0 \\
QuantVLA & \underline{98.7} & \underline{61.44} & 13.74 & \underline{174.0} & 98.0 & 66.18 & 15.08 & \underline{171.0} \\
HoloQ-VLA & 97.0 & 76.80 & 13.18 & 176.0 & \textbf{100.0} & 82.20 & \underline{14.21} & 174.0 \\
\runtimeBand
\textbf{\textit{\method{}}} & \textbf{98.8} & \textbf{53.58} & \textbf{11.45} & \textbf{145.0} & 98.4 & \textbf{55.85} & \textbf{13.18} & \textbf{143.0} \\
\specialrule{.3pt}{1.4pt}{1.0pt}
 & \multicolumn{4}{c}{\textbf{Goal}} & \multicolumn{4}{c}{\textbf{Long}} \\
\midrule
ActQuant & \underline{96.4} & 65.70 & \underline{15.18} & 181.0 & 91.6 & 93.85 & \underline{19.92} & 235.8 \\
QuantVLA & \textbf{98.0} & \underline{61.72} & 17.26 & \underline{177.0} & \textbf{96.0} & \underline{91.98} & 26.60 & \underline{224.5} \\
HoloQ-VLA & 96.0 & 80.40 & 15.62 & 179.0 & \underline{94.0} & 99.48 & 23.55 & 231.1 \\
\runtimeBand
\textbf{\textit{\method{}}} & \textbf{98.0} & \textbf{48.91} & \textbf{14.25} & \textbf{147.0} & \textbf{96.0} & \textbf{62.45} & \textbf{14.04} & \textbf{163.3} \\
\bottomrule
\end{tabular*}
\endgroup

\end{minipage}\hspace{2.5mm}%
\begin{minipage}[t]{54.1mm}
\setlength{\parskip}{0pt}
\vspace{0pt}
\centering
\begingroup
\setlength{\parskip}{0pt}
\setlength{\parindent}{0pt}
\definecolor{dyqPlotTranslation}{HTML}{4B83BA}
\definecolor{dyqPlotRotation}{HTML}{AD6129}
\definecolor{dyqPlotReference}{HTML}{959EA8}
\fontsize{7}{8.2}\selectfont
\noindent
\begin{minipage}[t]{.5\linewidth}
\centering
\textcolor{dyqPlotTranslation}{\textbf{Translation}}\\[-.4pt]
{\fontsize{6}{7}\selectfont Dynamic precision}
\end{minipage}%
\begin{minipage}[t]{.5\linewidth}
\centering
\textcolor{dyqPlotRotation}{\textbf{Rotation}}\\[-.4pt]
{\fontsize{6}{7}\selectfont Compensation}
\end{minipage}
\par\vspace{1.8pt}
{\fontsize{6.5}{7.5}\selectfont
\textcolor{dyqPlotReference}{\rule[1.6pt]{1.2mm}{.65pt}\kern.55mm
\rule[1.6pt]{1.2mm}{.65pt}\kern.55mm\rule[1.6pt]{1.2mm}{.65pt}}\kern1mm w/o
\hspace{3mm}
\textcolor{dyqPlotTranslation}{\rule[1.6pt]{2.2mm}{.65pt}}%
\textcolor{dyqPlotRotation}{\rule[1.6pt]{2.2mm}{.65pt}}\kern1mm w/
\par}
\vspace{1pt}
\includegraphics[width=\linewidth,trim=0 0 0 270bp,clip]{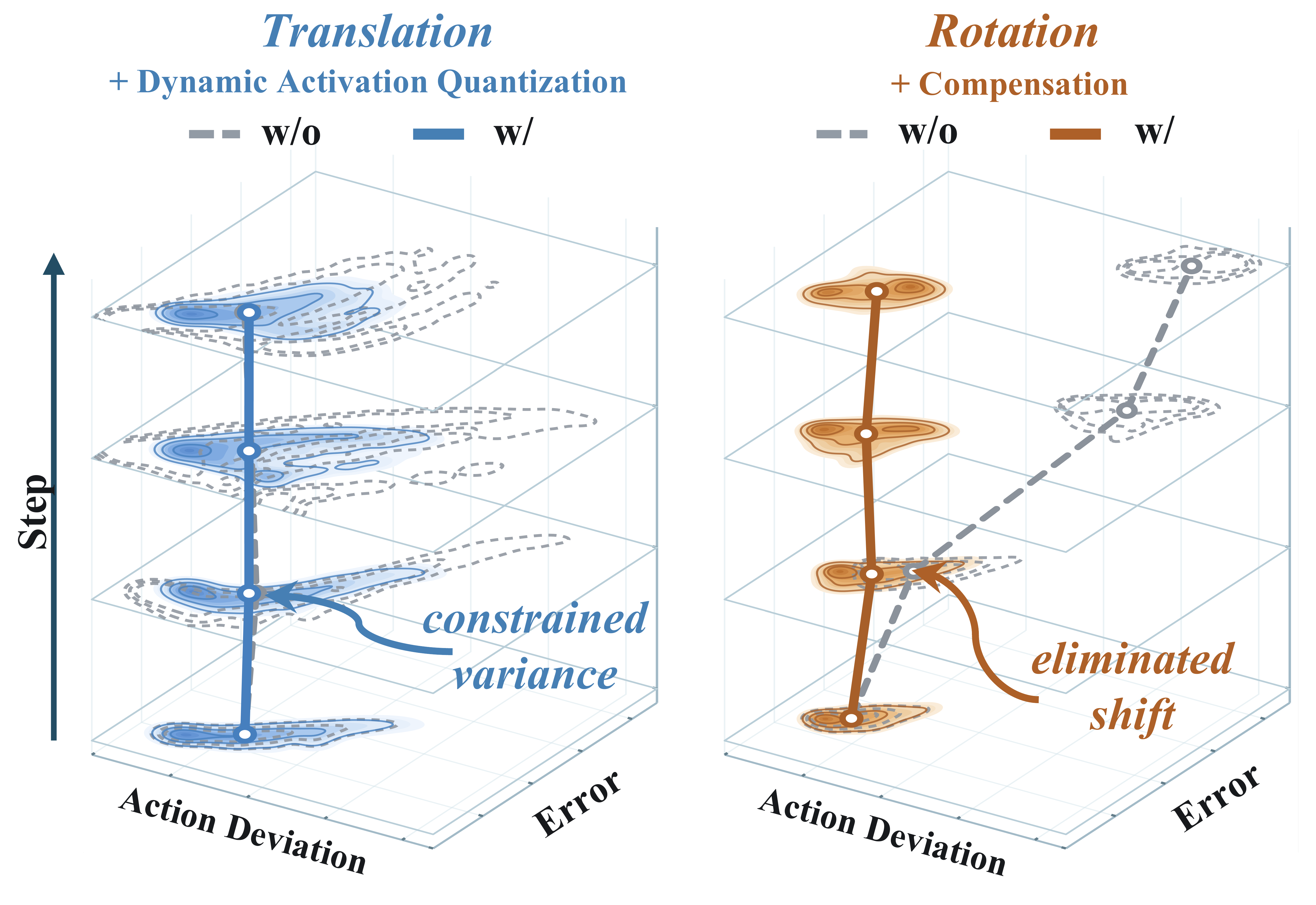}
\par
\endgroup

\captionsetup{position=bottom}
\captionof{figure}{Translational and Rotational Error Distribution Ablation.}
\label{fig:ablation_error_distribution}
\end{minipage}
\end{minipage}\par
\FloatBarrier

\FloatBarrier

\noindent
\begin{minipage}[t]{64.2mm}
\vspace{0pt}
\textbf{Ablation Study.}
Compensation (Comp.) raises W4A4 SR from 94.1\% to 96.3\% (\textcolor{orange}{Tab.~\ref{tab:ablation}}).
With dynamic precision, it raises SR from 97.2\% to 97.8\%, reduces mean steps from 155.1 to 149.5, and improves rollout speedup from 1.89$\times$ to 1.93$\times$.
Shuffling precision order ($A_{\mathrm{shuf}}$) with compensation increases errors and steps, reducing SR to 96.7\% and rollout speedup to 1.82$\times$.
\textcolor{orange}{Fig.~\ref{fig:ablation_error_distribution}} shows that dynamic precision constrains translation error variance, while compensation suppresses shifts in rotation error distributions across steps.
\end{minipage}\hspace{3mm}%
\begin{minipage}[t]{72.5mm}
\captionsetup{position=top}
\captionof{table}{Ablation on $\pi_{0.5}$ across LIBERO suites.
Rollout denotes measured speedup over BF16.}
\label{tab:ablation}
\centering
\begingroup
\fontsize{9}{10.2}\selectfont
\setlength{\tabcolsep}{0pt}
\renewcommand{\arraystretch}{1.035}
\setlength{\aboverulesep}{1pt}
\setlength{\belowrulesep}{1pt}
\setlength{\cmidrulesep}{0.5pt}
\setlength{\heavyrulewidth}{0.65pt}
\setlength{\lightrulewidth}{0.35pt}
\setlength{\cmidrulewidth}{0.25pt}
\providecolor{dyqMainBlue}{HTML}{F0F5FA}

\makeatletter
\def\ablationBand{%
  \noalign{%
    \begingroup
    \color{dyqMainBlue}
    \hrule height\dimexpr
      \ht\@arstrutbox+\dp\@arstrutbox
    \relax
    \endgroup
    \vskip-\dimexpr
      \ht\@arstrutbox+\dp\@arstrutbox
    \relax
  }%
}
\makeatother

\begin{tabular*}{\linewidth}{@{\extracolsep{\fill}}lcccccc@{}}
\toprule
\multirow{2}{*}{\textbf{Prec.}}
& \multirow{2}{*}{Comp.}
& \multirow{2}{*}{SR $\uparrow$}
& \multicolumn{2}{c}{Error $\downarrow$}
& \multirow{2}{*}{Steps $\downarrow$}
& \multirow{2}{*}{Rollout $\uparrow$}\\
\cmidrule(lr{0.8mm}){4-5}
& & & Trans. & Rot. & & \\
\midrule
BF16 ref. & ---
& 97.0 & 60.37 & 14.19 & 159.7 & 1.00\\
\midrule
W4A8 & $\times$
& 94.4 & 74.94 & 19.06 & 168.2 & 1.29\\
W4A4 & $\times$
& 94.1 & 91.47 & 20.75 & 171.5 & 1.44\\
W4A4 & $\checkmark$
& 96.3 & 68.72 & 15.81 & 160.8 & 1.49\\
W4/A$_{\mathrm{dyn}}$ & $\times$
& \underline{97.2} & \underline{61.84} & \underline{14.47}
& \underline{155.1} & \underline{1.89}\\
W4/A$_{\mathrm{shuf}}$ & $\checkmark$
& 96.7 & 65.84 & 15.26 & 158.4 & 1.82\\
\ablationBand
\textbf{W4/A}\textbf{\boldmath$_{\mathrm{dyn}}$} & $\checkmark$
& \textbf{97.8} & \textbf{55.20} & \textbf{13.23}
& \textbf{149.5} & \textbf{1.93}\\
\bottomrule
\end{tabular*}
\endgroup

\end{minipage}
\par

\section{Conclusion}
\label{tex:conclusion}
This paper analyzes the correlation between actions and errors in closed-loop VLA execution, revealing that quantization disperses the translation error distribution and induces cumulative drift in rotation errors. 
Based on this observation, we propose \method{}, which dynamically selects activation precision and compensates accumulated residuals to simultaneously control quantization overhead and runtime errors. 
We believe \method{} facilitates future VLA deployment.

\clearpage
\section*{AI Use Statement}
We use AI solely for language polishing and improving the clarity of the manuscript. 
All technical content, experimental design, analysis, and conclusions were developed and verified by the authors.

\bibliography{ref/reference}
\bibliographystyle{iclr2027_conference}

\appendix
\clearpage
\appendix

\section{Extended Related Work}
\label{app:extended_related_work}

\subsection{Efficient Inference for VLA Models.}

To reduce the computational and memory costs of VLA inference, existing studies explore three broad directions: (1) model compression, (2) runtime optimization, and (3) architectural design.

\textbf{Model Compression.} Compression methods reduce redundant computation in action prediction through visual token selection and model layer pruning~\citep{efficientvla,specprunevla,lightvla}.
For example, LightVLA learns token selection through differentiable pruning, retaining visual information useful for task execution~\citep{lightvla}.

\textbf{Runtime optimizations.} Runtime optimizations improve how computation is executed and reused.
Speculative decoding reduces sequential decoding costs by rapidly drafting action tokens and verifying them in parallel~\citep{specvla,kerv}.
Caching methods reuse visual computation across observations or intermediate features across diffusion denoising steps~\citep{vlacache,efficientvla}.
Early exiting and dynamic layer skipping further adapt the executed model depth to computational budgets or inputs~\citep{deervla,molevla}.

\textbf{Architecture Design.} 
Architectural approches reduce model costs through compact multimodal backbones and dedicated action modules, as exemplified by TinyVLA~\citep{tinyvla} and SmolVLA~\citep{smolvla}.
Action representations and prediction mechanisms are also redesigned: FAST compresses action token sequences~\citep{fast}, while OpenVLA-OFT combines action chunking with parallel decoding to accelerate action generation~\citep{openvla_oft}.

\subsection{Quantization of Vision-Language-Action Models}

To improve the accuracy-speed trade-off of low-precision VLA inference, existing studies explore three directions: (1) distribution-aware quantization, (2) sensitivity-aware quantization, and (3) trajectory-aware quantization.

\textbf{Distribution-Aware Quantization.}
Distribution-aware methods mitigate quantization distortion through representation adaptation and numerical calibration.
BitVLA~\citep{bitvla} combines quantization with distillation to preserve model capabilities, while QuantVLA~\citep{quantvla} calibrates attention temperatures and residual output scales.
HoloQ-VLA~\citep{holoqvla} addresses weight and activation distortion through composite rotations and denoising-step-dependent scaling.
SQAP-VLA~\citep{sqap} further coordinates quantization with visual token pruning, accounting for the effects of quantized representations on token selection.

\textbf{Sensitivity-Aware Quantization.}
Sensitivity-aware methods assign different bit widths to model components according to their influence on action prediction.
QVLA~\citep{qvla} uses action output sensitivity to guide channel-wise bit allocation and validates this sensitivity measure over short horizons.
ActQuant~\citep{actquant} similarly allocates precision according to action relevance, assigning a bit width to each weight matrix based on its contribution to action prediction.

\textbf{Trajectory-Aware Quantization.}
Trajectory-aware methods further incorporate trajectory drift and execution-stage information into offline precision allocation.
DA-PTQ~\citep{daptq} combines representation compensation with drift-aware layerwise bit allocation, using offline estimates of trajectory drift to identify components that require higher precision.
Mix-QVLA~\citep{mixqvla} uses task evidence and phase-dependent sensitivity to guide layerwise precision selection, accounting for changes in layer importance across task execution.
Both methods use trajectory or temporal information during calibration to determine precision configurations that remain fixed during deployment.

\subsection{Core Differences Between \textit{DyQ-VLA} and Prior Works}
\textbf{SOTA Quantization Works.} 
Prior studies follow conventional approaches to large-model quantization, focusing primarily on precision selection and the accuracy-speed trade-off.
Although some methods consider the closed-loop nature of VLA execution, how runtime errors evolve and how quantization alters the action-error correlation remain insufficiently characterized.

\textbf{\textit{DyQ-VLA}'s Key Innovation.}
In contrast, we explicitly analyze the action-error correlation, reveal its distinct characteristics in the translational and rotational components, and characterize the impact of quantization on this correlation.
Building on these findings, we rethink quantization strategy design for VLA models and propose \textit{DyQ-VLA}, a runtime-error-aware dynamic quantization framework.


\section{Definitions and Measurement of Runtime Error}
\label{app:error_measurement}

\subsection{Sources and Measurement of Runtime Error}
\label{app:runtime_error_measurement}

Runtime error can arise from imperfect action prediction.
VLA models learn a mapping from observations to actions using
finite demonstration data.
Limited data coverage and imperfect model fitting can cause
predicted translation and rotation actions to differ from
reference actions during deployment.
Executing these actions can therefore produce poses that
deviate from the reference trajectory.
Finite controller tracking accuracy and contact constraints
can also cause actual motion to differ from commanded motion.
These deviations alter subsequent observations, potentially
leading to states poorly covered by demonstrations and
affecting later action predictions.
Quantization approximates weights and activations, further
altering action outputs and the evolution of execution deviations.
We measure runtime error through the discrepancy between
the actual and reference poses of the end effector.
We measure this deviation using end-effector poses after each
execution chunk.
Let $t$ index policy queries, with the first $h$ predicted actions
executed before the next query.
Denote the resulting position and unit quaternion by
$\mathbf p_t$ and $\mathbf q_t$, and the corresponding reference
pose in the same coordinate frame by
$\bar{\mathbf p}_t$ and $\bar{\mathbf q}_t$.
Translation and rotation errors are defined as
\begin{equation}
\begin{aligned}
e_t^T
&= \|\mathbf p_t-\bar{\mathbf p}_t\|_2,\\
e_t^R
&= 2\arccos\!\left(
\min\!\left\{1,\left|\mathbf q_t^\top\bar{\mathbf q}_t\right|\right\}
\right).
\end{aligned}
\label{eq:app_execution_error}
\end{equation}
The rotation error is the shortest angle between the two
orientations.
The absolute value accounts for quaternion sign ambiguity,
while clipping prevents numerical round-off from producing
invalid inputs to $\arccos$.
Translation and rotation errors are computed in meters and
radians and reported in millimeters and degrees, respectively.

On LIBERO, $h=5$ for $\pi_{0.5}$ and $h=1$ for GR00T-N1.5.
The query index $t$ is distinct from control-step and internal
flow-step indices.
For action--error analysis, the error measured after executing
the actions from query $t$ is paired with the action predictions
at query $t+1$.

\subsection{Reference Trajectories}
\label{app:reference_trajectories}
Task evaluation uses matched and aligned demonstration trajectories as references for both BF16 and quantized models. Success rate is computed over all episodes, whereas runtime errors and execution steps are averaged over successful episodes. 
In the LIBERO main table, errors are averaged equally across the four suite-level means, while execution steps are aggregated using the number of successful episodes in each suite as weights. 
Action--Error Correlation and sensitivity analyses use unperturbed rollouts at the same precision as references. 
The reference and perturbed branches share the same initial state and matched policy randomness, and each proceeds in a closed loop using its own observations after the perturbation. 
We pair pose deviations at corresponding execution steps with action-prediction differences at the next policy query, and characterize the effects of quantization through changes in the joint distributions and sensitivities across precision settings.

\subsection{Action-Error Joint Distributions}
\label{app:joint_distributions}

Following the paired rollout protocol in \textcolor{orange}{Appendix~\ref{app:reference_trajectories}}, we inject
action perturbations of equal magnitude at a fixed policy query
and record subsequent execution errors and differences in
action predictions. The normalized error after executing
query $t$ and action discrepancy at the next query are
\begin{equation}
\begin{aligned}
\varepsilon_t
&=
\sqrt{
\left(\frac{e_t^T}{s_T}\right)^2+
\left(\frac{e_t^R}{s_R}\right)^2
},\\
\Delta A_{t+1}
&=
\sqrt{
\frac{1}{h}\sum_{i=1}^{h}\sum_{d=1}^{6}
\left(
\frac{a_{t+1,i,d}-\bar a_{t+1,i,d}}{s_d^A}
\right)^2
}.
\end{aligned}
\label{eq:app_joint_variables}
\end{equation}
Here, $a$ and $\bar a$ denote commands for the end effector
passed to the LIBERO controller by the perturbed and reference
branches, respectively, both operating at the same precision.
The six channels comprise three translation and three rotation
components, excluding the gripper.
The scales $s_T$, $s_R$, and $s_d^A$ are calibrated as described
in \textcolor{orange}{Appendix~\ref{app:scale_calibration}} and held fixed across comparisons.
Each paired rollout contributes a sample
$(\Delta A_{t+1},\varepsilon_t)$ at query $t$.
For temporal comparisons, we retain a fixed set of $N$ paired
rollouts with valid records at every displayed checkpoint.

At a fixed query, let
$x_n=\Delta A_{t+1}^{(n)}$ and $y_n=\varepsilon_t^{(n)}$.
We estimate their joint density using Gaussian kernels
with reflection at the zero boundaries:
\begin{equation}
\hat p_t(x,y)
=
\frac{1}{N}\sum_{n=1}^{N}
\bigl[K_x(x-x_n)+K_x(x+x_n)\bigr]
\bigl[K_y(y-y_n)+K_y(y+y_n)\bigr],
\qquad x,y\geq 0,
\label{eq:app_joint_density}
\end{equation}
where
\begin{equation*}
K_x(u)=
\frac{1}{\sqrt{2\pi}\sigma_x}
\exp\!\left(-\frac{u^2}{2\sigma_x^2}\right),
\end{equation*}
and $K_y$ is defined analogously.
Each bandwidth is the corresponding coordinate's standard
deviation pooled across checkpoints, multiplied by $N^{-1/6}$,
with a lower bound equal to the larger of $2\%$ of its pooled
maximum and $10^{-4}$.
Bandwidths remain fixed across checkpoints for each component.
Distribution centers and dispersion are measured by the
means and variances of the original samples.

We further extract translation and rotation components from
the same paired rollouts. Let $D_T=\{1,2,3\}$ and
$D_R=\{4,5,6\}$ denote the corresponding action channels.
For $X\in\{T,R\}$, we define
\begin{equation}
\varepsilon_t^X=\frac{e_t^X}{s_X},
\qquad
\Delta A_{t+1}^X
=
\sqrt{
\frac{1}{h}\sum_{i=1}^{h}\sum_{d\in D_X}
\left(
\frac{a_{t+1,i,d}-\bar a_{t+1,i,d}}{s_d^A}
\right)^2
}.
\label{eq:app_component_pairs}
\end{equation}
Applying the same density estimator to
$(\Delta A_{t+1}^T,\varepsilon_t^T)$ and
$(\Delta A_{t+1}^R,\varepsilon_t^R)$ yields the translation
and rotation action-error distributions, respectively,
for comparing their centers and dispersion across execution.

\subsection{Signed Rotational Residuals}
\label{app:signed_rotational_residuals}

Let $\mathbf x_{k,n,i}$ denote the monitored activation for
action $i$ at flow step $n$ of query $k$.
The signed quantization residual is
\begin{equation}
\boldsymbol{\delta}_{k,n,i}
=
\operatorname{QDQ}(\mathbf x_{k,n,i})
-
\mathbf x_{k,n,i},
\label{eq:app_activation_residual}
\end{equation}
where $\operatorname{QDQ}$ denotes quantization followed by
dequantization at the current precision.
We project these residuals onto rotational components
and accumulate them across queries:
\begin{equation}
\boldsymbol{\zeta}_k^R
=
\overline{\mathbf B}_R
\sum_{n=1}^{N_f}\Delta\tau_n
\sum_{i=1}^{h}\boldsymbol{\delta}_{k,n,i},
\qquad
\boldsymbol{\mathcal R}_t^R
=
\sum_{k=0}^{t-1}\boldsymbol{\zeta}_k^R.
\label{eq:app_rotation_projection}
\end{equation}
Here, $N_f$ is the number of flow steps and $\Delta\tau_n$
is the sampler's signed integration coefficient.
The fixed matrix $\overline{\mathbf B}_R$ averages local
Jacobians from monitored activations to rotational flow
outputs over calibration samples and flow steps.
Accumulation uses a common coordinate frame and fixed scales,
preserving reinforcement and cancellation between residuals.
For error tracking, we pair $e_t^R$, measured after executing
query $t$'s actions, with $\|\boldsymbol{\mathcal R}_{t+1}^R\|_2$,
which includes the current query's contribution.

\subsection{Response Sensitivity}
\label{app:response_sensitivity}

At each precision, response sensitivity measures the change
in subsequent action predictions per unit normalized execution
error. For paired samples with perturbation direction $j$,
\begin{equation}
\mathscr S_{t,j}^{X}
=
\frac{\Delta A_{t+1}}{\varepsilon_t^{X}},
\qquad
X\in\{T,R\},\quad \varepsilon_t^{X}>0.
\label{eq:app_response_sensitivity}
\end{equation}
Here, $\Delta A_{t+1}$ and $\varepsilon_t^X$ follow
\textcolor{orange}{Appendix~\ref{app:joint_distributions}}. The numerator includes both translation and
rotation action changes to capture responses across components,
while the denominator uses the corresponding normalized error.
We compare FP and quantized sensitivities by query and
perturbation direction under matched tasks, initial states,
and perturbation conditions, using identical normalization
scales to characterize quantization effects on action-error
responses.

\subsection{Calibration of Scales}
\label{app:scale_calibration}

We calibrate scales using 512 FP LIBERO rollouts per backbone,
initialized from training demonstrations without injected
perturbations, approximately balanced across analyzed tasks,
and separate from evaluation.
For each execution chunk $c$ in the calibration set $\mathcal C$,
let $\Delta p_c$ and $\Delta\theta_c$ denote the translation and
rotation magnitudes between its initial and final poses,
computed using \textcolor{orange}{Eq.~(\ref{eq:app_execution_error})}.
With $Q_\alpha$ denoting the empirical $\alpha$ quantile,
we define: 
\begin{equation}
\begin{aligned}
s_T
&= Q_{0.95}\!\left(\{\Delta p_c:c\in\mathcal C\}\right),\\
s_R
&= Q_{0.95}\!\left(\{\Delta\theta_c:c\in\mathcal C\}\right),\\
s_d^A
&= \frac{Q_{0.99}(a_d)-Q_{0.01}(a_d)}{2}.
\end{aligned}
\label{eq:app_calibration_scales}
\end{equation}
Here, $a_d$ contains the executed values of action channel $d$,
whose scale follows percentile normalization~\citep{openvla,fast}.
For each backbone, scales remain fixed across precision settings,
queries, and perturbation conditions.
Normalized differences are not clipped.


\section{Additional Runtime Error Analysis}
\label{app:runtime_error_analysis}
\subsection{Trajectory Visualizations.}
\begin{figure}[!b]
    \centering
    \includegraphics[width=5.5in]{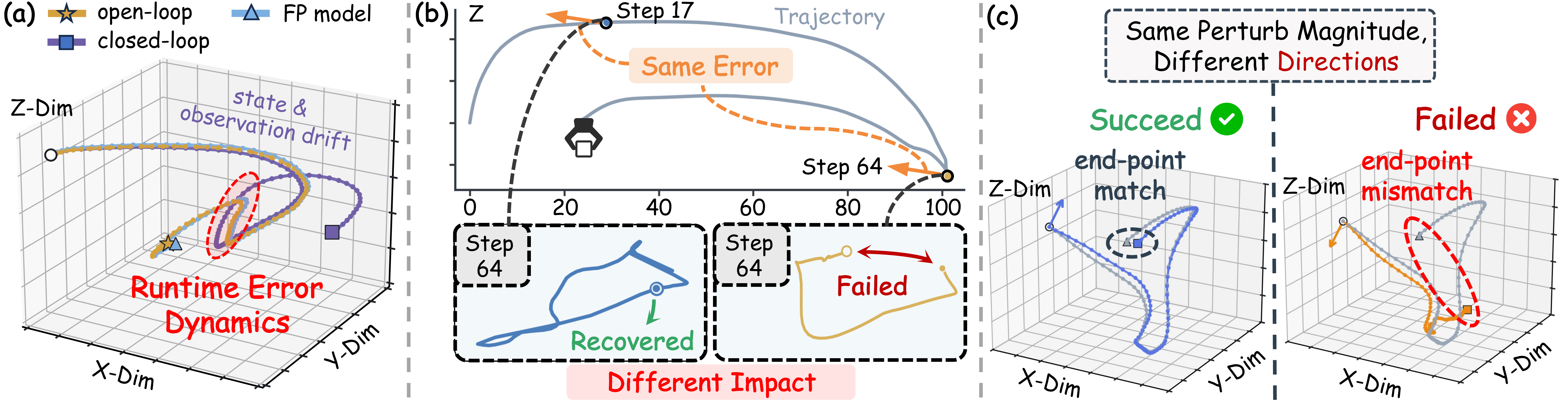}
    \caption{
    \textbf{Trajectory Visualizations of Runtime Error Dynamics and Their Behavioral Impact.}
    (a)~FP reference, open-loop evaluation, and closed-loop execution trajectories.
    (b)~Identical action offsets at different steps, with trajectories projected onto the $yz$ plane.
    (c)~Offsets of equal magnitude in different directions at the same step.
    }
\label{fig:analysis_case}
\end{figure}

We provide trajectory visualizations to complement the runtime error analysis in \textcolor{orange}{Sec.~\ref{sec:analysis}}.
\textcolor{orange}{Fig.~\ref{fig:analysis_case}~(a)} compares the FP reference, open-loop evaluation, and closed-loop execution trajectories, illustrating how action errors evolve through feedback between the policy and its environment.
In \textcolor{orange}{Fig.~\ref{fig:analysis_case}~(b)}, the same action offset is injected at different execution steps, producing distinct trajectories: one rollout recovers from the resulting error, while the other fails.
\textcolor{orange}{Fig.~\ref{fig:analysis_case}~(c)} fixes the injection step and offset magnitude while varying the direction, yielding different trajectories and task outcomes.
Even offsets of the same magnitude affect subsequent behavior differently depending on the execution state and offset direction.

\subsection{Action-Error Distributions Across Models and Benchmarks}
\label{app:cross_setting_distributions}

We extend the analysis in Sec.~\ref{sec:analysis_1}
to GR00T-N1.5 on the four LIBERO suites and $\pi_{0.5}$
on RoboTwin 2.0 under Clean and Randomized settings,
examining whether the component differences persist across
models and environments under full precision.
Following the paired rollout and density estimation procedures
in Sec.~\ref{app:joint_distributions}, we construct joint
distributions of runtime error and subsequent action discrepancy
for translation and rotation, and compare their evolution
across policy queries.

Fig.~\ref{fig:app_tr_distributions} shows a consistent pattern
across all six settings.
Translation distributions remain relatively dispersed around
stable centers, whereas rotation distributions are more
concentrated and their centers progressively shift toward
larger errors.
These results support Observation~\ding{172} in
Sec.~\ref{sec:analysis_1} across model backbones, task types,
and benchmark settings.

\begin{figure}[t]
    \centering
    \includegraphics[width=\linewidth]{fig/fig_distributions_six.pdf}
    \caption{\textbf{Action-Error Distributions Across Models and Benchmarks.}}
    \label{fig:app_tr_distributions}
\end{figure}

\begin{figure}[!b]
    \centering
    \includegraphics[width=\linewidth]{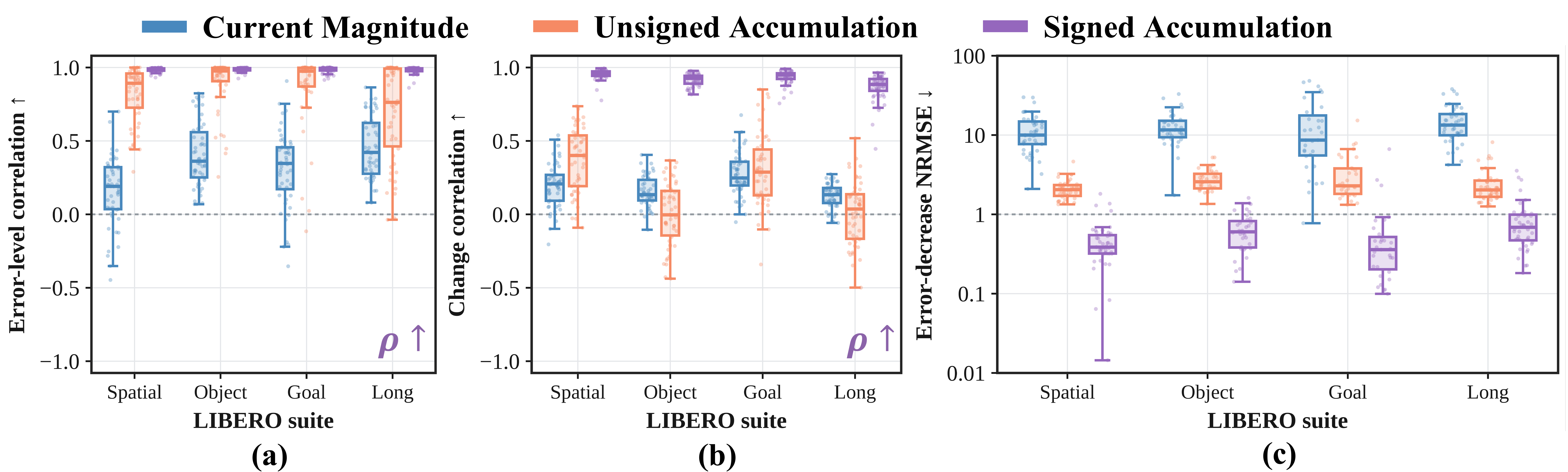}
    \caption{
    \textbf{Tracking Rotational Runtime Error.}
    (a) Spearman correlation with rotational runtime error.
    (b) Spearman correlation between signal changes and error changes.
    (c) NRMSE of normalized changes for error decreases $\ge0.5^\circ$;
    the dashed line marks the zero-change baseline.
    Dots denote trajectories; boxes show medians and interquartile
    ranges, with $1.5\,\mathrm{IQR}$ whiskers.
    }
    \label{fig:cross_task}
\end{figure}

\subsection{Tracking Runtime Error with Quantization Residuals}
\label{app:residual_tracking}

We evaluate whether quantization residuals track rotational
runtime error using 200 trajectories across four LIBERO suites,
with 10 tasks per suite and five initial states per task.
We compare current residual magnitude, unsigned accumulation,
and signed accumulation, aligned with the rotational error
$e_t^R$ measured after executing query $t$.
For each trajectory, we compute Spearman correlations with
error levels and with error changes between successive queries.
Fig.~\ref{fig:cross_task}(a) shows that unsigned and signed
accumulation both correlate strongly with error levels.
However, unsigned accumulation tracks error changes less
consistently, with median change correlations near zero on
Object and Long, whereas signed accumulation maintains high
correlations across all suites (Fig.~\ref{fig:cross_task}(b)).

This difference follows from how residual history is aggregated.
Current magnitude ignores historical contributions, while
unsigned accumulation discards directional information.
Let $u_t=\sum_{k=0}^{t}\|\boldsymbol{\zeta}_k^R\|_2$ denote
unsigned accumulation and
$s_t=\|\boldsymbol{\mathcal R}_{t+1}^R\|_2$ denote the
magnitude of signed accumulation, both including query $t$.
The residual definition in Eq.~(\ref{eq:app_rotation_projection})
gives
\begin{equation}
\begin{aligned}
u_t-u_{t-1}
&=\|\boldsymbol{\zeta}_t^R\|_2\ge0,\\
s_t^2-s_{t-1}^2
&=\|\boldsymbol{\zeta}_t^R\|_2^2
+2\left\langle
\boldsymbol{\mathcal R}_t^R,\boldsymbol{\zeta}_t^R
\right\rangle.
\end{aligned}
\label{eq:app_residual_magnitude_change}
\end{equation}
Unsigned accumulation cannot decrease, even when it correlates
strongly with overall error levels.
Signed accumulation retains the interaction between current
and historical residuals: aligned contributions reinforce
each other, while sufficiently strong opposing contributions
reduce the accumulated magnitude.
It therefore captures both reinforcement and cancellation
across queries.

We further test whether this ability to decrease corresponds
to observed error reductions.
Each signal and $e_t^R$ is divided by its own trajectory maximum.
At checkpoints where the unnormalized rotation error decreases
by at least $0.5^\circ$, NRMSE is the RMS difference between
normalized signal and error changes, divided by the RMS
normalized error change.
Here, normalized error changes are negative, whereas changes
in unsigned accumulation remain nonnegative.
Consequently, $\mathrm{NRMSE}_u\ge1$, where one is the baseline
for predicting zero change.
In Fig.~\ref{fig:cross_task}(c), signed accumulation achieves
median NRMSE below this baseline in every suite, showing that
it follows decreases in measured rotational runtime error.

\subsection{Feedback Estimation of Response Sensitivity}
\label{app:feedback_sensitivity}

\noindent
\begin{minipage}[t]{0.46\textwidth}
    \vspace{0pt}
Fig.~\ref{fig:kinematics} shows that feedback estimates track
the relative strength of translation response sensitivity
across both backbones and all four LIBERO suites.
Median Spearman correlations range from $0.83$ to $0.92$
across queries (Temporal) and from $0.77$ to $0.87$ across
perturbation directions (Directional).
We evaluate 40 trajectories per backbone and suite,
for 320 trajectories in total.
Measured sensitivity follows
Eq.~(\ref{eq:app_response_sensitivity}) with $X=T$:
the subsequent action response, including translation and
rotation components, per unit normalized translation error.
Temporal correlations hold the perturbation direction fixed,
while Directional correlations hold the query fixed.
These results show that the estimates distinguish sensitivity
across both execution stages and perturbation directions.
\end{minipage}
\hfill
\begin{minipage}[t]{0.5\textwidth}
    \vspace{0pt}
    \centering
    \includegraphics[width=\linewidth,keepaspectratio]{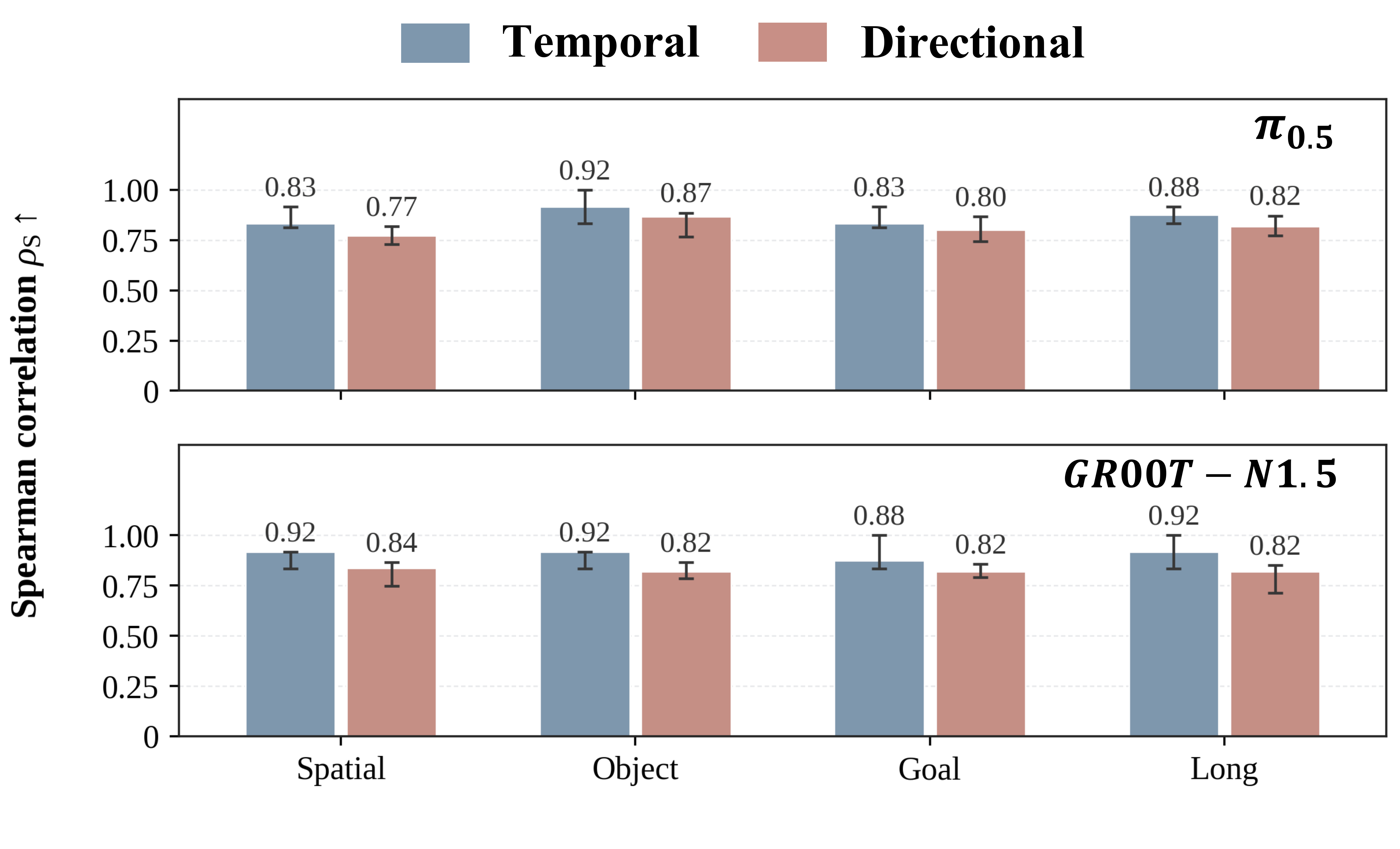}
    \captionof{figure}{
    \textbf{Feedback Sensitivity Estimation.}
    Spearman correlations with measured translation response
    sensitivity across queries (Temporal) and perturbation
    directions (Directional).
    Bars show medians across 40 trajectories per backbone and
    LIBERO suite; error bars span the 25th to 75th percentiles.
    }
    \label{fig:kinematics}
\end{minipage}
\par\medskip

The controller derives feedback scores from executed motion
and the remaining action plan.
Let $\mathbf a_{t,i}^{T}$ and $\mathbf a_{t,i}^{R}$ denote
translation and rotation commands at query $t$.
After executing the first $h$ actions,
\begin{equation}
\begin{aligned}
\mathbf c_t &= \sum_{i=1}^{h}\mathbf a_{t,i}^{T},
& \mathbf d_t &= \mathbf p_t-\mathbf p_{t-1},\\
g_t &= \frac{\mathbf d_t^{\top}\mathbf c_t}
{\|\mathbf c_t\|_2^2+\epsilon},
& G_t &= \max\{G_{t-1},g_t,\epsilon\}.
\end{aligned}
\label{eq:app_feedback_gain}
\end{equation}
Here, $g_t$ measures execution gain along the commanded
direction, $G_t$ is its running maximum initialized to zero,
and $\epsilon=10^{-12}$.
The normalized scores are
\begin{equation}
\begin{aligned}
\widehat{s}_{t+1,T}
&=\frac{[1-g_t/(G_t+\epsilon)]_+}{\theta_T},\\
\widehat{s}_{t+1,R}
&=\frac{1}{\theta_R(h-1)}
\sum_{i=h+1}^{2h-1}
\|\mathbf a_{t,i+1}^{R}-\mathbf a_{t,i}^{R}\|_2.
\end{aligned}
\label{eq:app_feedback_scores}
\end{equation}
Here, $[x]_+=\max(x,0)$ and $\theta_T,\theta_R>0$
are fixed normalization thresholds.
Lower execution gain and greater rotation variation in the
unexecuted tail increase the respective scores, tightening
the constraints in Eq.~(\ref{eq:4-1}) without additional
policy evaluations.

For correlation analysis, the feedback estimate
$\widehat{\mathscr S}_{t,j}^{T}$ formed after query $t$
is compared with the measured sensitivity
$\mathscr S_{t,j}^{T}$ of the subsequent action response:
\begin{equation}
\begin{aligned}
\rho_{\mathrm{Temporal}}(j)
&=\operatorname{Spearman}_{t}
\left(\widehat{\mathscr S}_{t,j}^{T},\mathscr S_{t,j}^{T}\right),\\
\rho_{\mathrm{Directional}}(t)
&=\operatorname{Spearman}_{j}
\left(\widehat{\mathscr S}_{t,j}^{T},\mathscr S_{t,j}^{T}\right).
\end{aligned}
\label{eq:app_feedback_correlations}
\end{equation}
The correlations support feedback-based sensitivity ordering
across queries and perturbation directions for dynamic
precision selection (Sec.~\ref{sec:method_1}).

\section{Method and Implementation Details}
\label{app:method_details}


\subsection{Algorithm Overview}
\label{app:runtime_algorithm}

Algorithm~\ref{alg:dyq_vla} summarizes the offline preparation
and online execution of \method{}.
Offline, we quantize the target weights to 4 bits, calibrate
the rotational residual projection, and prepare the execution
paths for candidate activation configurations.
We profile their error estimates $E_d(b)$ and inference
latencies $L(b)$, and calibrate the thresholds $\tau_d$
using the reference configuration.
These quantities remain fixed during online execution.

Before each policy query, the controller selects a configuration
using the sensitivity estimated from previous execution feedback
and the residual remaining after compensation.
Following Eq.~(\ref{eq:4-1}), it chooses the lowest-latency
configuration that passes the calibrated test, or uses the
predefined fallback if none passes.
The selected path produces an action chunk and activation
residual records $\mathcal Q_t$.
Their rotational projection is combined with the historical
state $\mathcal H_{t-1}$ to obtain $\widehat{\mathbf e}_t$.
The rotation commands are then corrected according to
Eq.~(\ref{eq:4-2}).

After execution, we update the signed residual state using
the actual command corrections.
Both the residual contribution and the correction are computed
over the executed prefix of length $m_t\leq h$, accounting
for early termination.
The map $\rho_R$ converts the updated state to the scalar
residual scale used by the selection criterion, while $\Psi$
estimates the next query's sensitivity from execution feedback.
This completes the connection between precision selection,
rotation compensation, and subsequent execution.

\begin{algorithm}[t]
\caption{Runtime Error Control with \method{}}
\label{alg:dyq_vla}
\small
\begin{algorithmic}[1]

\REQUIRE Model $\pi$, instruction $\ell$,
calibration data $\mathcal D_{\mathrm{cal}}$;
candidate configurations $\mathcal B$,
reference $b_{\mathrm{ref}}$,
fallback $b_{\mathrm{fb}}\in\mathcal B$;
execution horizon $h$, gain $\eta\in(0,1]$;
calibrated sensitivity estimator $\Psi$,
residual-scale map $\rho_R$,
and initial sensitivity $\widehat{\mathbf s}_1\geq\mathbf 0$.

\ENSURE Action chunks executed with dynamic precision
and rotation compensation.

\STATE \textbf{Offline preparation}

\STATE $\mathbf W_4
    \gets \textsc{QuantizeWeights}
    (\pi,\mathcal D_{\mathrm{cal}})$

\STATE $\overline{\mathbf B}_R
    \gets \textsc{CalibrateProjection}
    (\pi,\mathcal D_{\mathrm{cal}})$

\STATE $\{\mathcal G_b\}_{b\in\mathcal B}
    \gets \textsc{PrepareRuntime}
    (\pi,\mathbf W_4,\mathcal B,\mathcal D_{\mathrm{cal}})$

\STATE Profile $E_d(b)\geq0$ and $L(b)$
    for $b\in\mathcal B$ and $d\in\{T,R\}$.

\STATE Calibrate positive thresholds $\tau_d$
    using $b_{\mathrm{ref}}$.

\STATE \textbf{Online execution}

\STATE $\mathcal H_0\gets\mathbf 0$,
    $r_{0,T}\gets0$, $r_{0,R}\gets0$, $t\gets1$

\WHILE{the task is not terminated}

    \STATE $o_t\gets\textsc{Observe}()$

    \STATE $\displaystyle
    \mathcal B_t\gets
    \left\{
    b\in\mathcal B:
    \max_{d\in\{T,R\}}
    \frac{
        \widehat s_{t,d}
        \bigl(E_d(b)+r_{t-1,d}\bigr)
    }{\tau_d}
    \leq1
    \right\}$

    \IF{$\mathcal B_t\neq\varnothing$}
        \STATE $b_t\gets
            \arg\min_{b\in\mathcal B_t}L(b)$
    \ELSE
        \STATE $b_t\gets b_{\mathrm{fb}}$
        \COMMENT{No candidate passed the calibrated test.}
    \ENDIF

    \STATE $(\mathbf A_t,\mathcal Q_t)
        \gets \textsc{Replay}
        (\mathcal G_{b_t},o_t,\ell)$

    \STATE $\widehat{\mathbf e}_t
        \gets \mathcal H_{t-1}
        +\textsc{ProjectResidual}
        (\mathcal Q_t,\overline{\mathbf B}_R,h)$

    \STATE $\widetilde{\mathbf A}_{t,1:h}
        \gets \textsc{CompensateRotation}
        (\mathbf A_{t,1:h},\widehat{\mathbf e}_t,\eta)$
        using \textcolor{orange}{Eq.~(\ref{eq:4-2})}.

    \STATE $(\mathcal F_t,
        \mathbf A^{\mathrm{appl}}_{t,1:m_t},m_t)
        \gets \textsc{ExecuteAndRecord}
        (\widetilde{\mathbf A}_{t,1:h})$

    \STATE $\displaystyle
        \Delta\mathbf a_t^{R,\mathrm{appl}}
        \gets
        \sum_{i=1}^{m_t}
        \left(
            \mathbf a_{t,i}^{R,\mathrm{appl}}
            -\mathbf a_{t,i}^{R}
        \right)$

    \STATE $\mathcal H_t
        \gets \mathcal H_{t-1}
        +\textsc{ProjectResidual}
        (\mathcal Q_t,\overline{\mathbf B}_R,m_t)
        +\Delta\mathbf a_t^{R,\mathrm{appl}}$

    \STATE $r_{t,R}\gets\rho_R(\mathcal H_t)$,
        $r_{t,T}\gets0$

    \STATE $\widehat{\mathbf s}_{t+1}
        \gets\Psi(\mathcal F_t,\mathbf A_t)$

    \STATE $t\gets t+1$

\ENDWHILE
\end{algorithmic}
\end{algorithm}

\begin{figure}[t]
    \centering
    \includegraphics[width=5.5in]{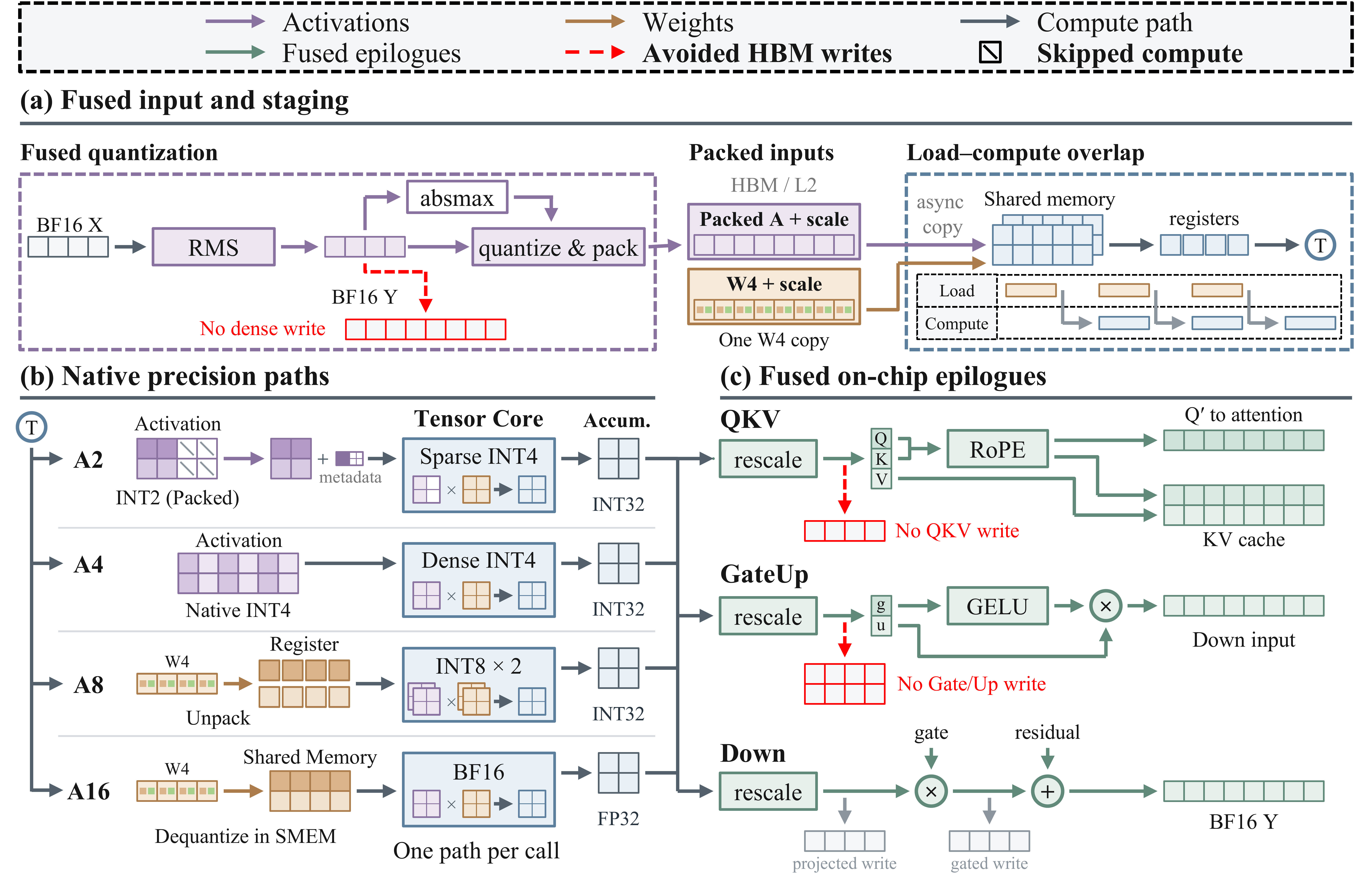}
    \caption{
    \textbf{Implementation details of the Proposed \method{} Framework.}
    }
    \label{fig:impl}
\end{figure}

\subsection{Quantized Kernels with Shared Weights}
\label{app:quantized_kernels}

Fig.~\ref{fig:impl} shows how the A2, A4, A8, and A16
paths share W4 storage on NVIDIA SM80.
Quantized weights are packed as two signed INT4 codes
per byte, with FP32 scales per output channel.
Modules access a single device allocation through views,
while each kernel converts its weight tiles into the
required operand format on chip.

The A2 path quantizes activations to ternary values and
retains two adjacent pairs within each group of eight.
The selected values and position metadata form a pairwise
4:8 sparse INT4 operand for \texttt{mma.sp}.
The A4 path directly uses dense INT4 multiplication.
The general A8 path expands W4 fragments to signed INT8
in registers and evaluates each $K=64$ slice with two
INT8 matrix operations.
The A16 path reconstructs BF16 weight tiles in shared
memory and uses BF16 multiplication with FP32 accumulation.
These conversions remain local to the current tile,
avoiding additional resident weight copies.

The integer paths accumulate into INT32 registers and
apply activation-row and weight-column scales before
BF16 output rounding.
The A16 path follows its corresponding weight-scaling
and rounding convention.
Thus, activation precision can change while the resident
weights remain in W4 format.

\subsection{Kernel Fusion and Runtime Scheduling}
\label{app:runtime_scheduling}

As illustrated in Fig.~\ref{fig:impl}, we fuse activation
preparation and output processing to reduce intermediate
memory traffic.
The activation producer combines normalization, affine
transformation, row-scale computation, and packing.
Normalized values remain in registers, and the kernel
writes packed activations and their scales for the
subsequent GEMM.

GEMM epilogues preserve the model's operation order and
BF16 rounding boundaries.
The QKV epilogue applies RoPE to Q and K and writes K/V
directly to the cache.
The GateUp epilogue combines the model's tanh-based GELU
with multiplication, producing a compact product instead
of separate Gate and Up tensors.
O and Down epilogues combine scaling, gating, and residual
addition into the final output store.
Where supported, Down quantizes its input during tile
loading, eliminating an additional packed-activation
write and read.

We specialize scheduling for the different Prefix and
Flow matrix shapes.
Asynchronous copies and multiple shared-memory stages
overlap tile loading with computation.
For the small-row Flow Down projection, split-K execution
increases parallelism, with INT32 partial sums stored in
a preallocated workspace and reduced in a fixed order
before output processing.

For the fixed $\pi_{0.5}$ deployment, a Prefix graph and
a Flow graph containing ten unrolled flow steps execute
through CUDA Graph replay.
Inputs, caches, outputs, and workspaces retain fixed
addresses.
Precision paths are prepared before execution, removing
allocation, weight repacking, compilation, and backend
search from replay.

\section{Extended Experiments}
\label{app:extended_experiments}

\subsection{Experimental Setup}
\textbf{Models and Benchmarks.}
We evaluate $\pi_{0.5}$~\citep{pi_0.5} and GR00T-N1.5~\citep{gr00t} on the Spatial, Object, Goal, and Long suites of LIBERO~\citep{libero}.
We additionally evaluate $\pi_{0.5}$ on RoboTwin 2.0~\citep{robotwin} under Clean and Randomized settings.
We use official checkpoints and benchmark protocols, with 50 evaluation rollouts per LIBERO task and 1000 rollouts per suite.
Inference runs on an NVIDIA A100 GPU.

\textbf{Baselines and Calibration.}
BF16 serves as the full-precision reference.
On LIBERO, we compare five general-purpose post-training quantization methods~\citep{gptq,awq,omniquant,smoothquant,duquant} and four VLA-specific methods~\citep{actquant,qvla,quantvla,holoqvla}; the latter are also evaluated on RoboTwin 2.0.
All quantized methods share 60 LIBERO calibration trajectories, which are used to configure quantization rather than counted as evaluation rollouts.
Baselines retain their original configurations, while \textit{DyQ-VLA} uses 4-bit weights and selects activation precision from $\{2,4,8,16\}$ bits during execution.

\subsection{Evaluation Metrics}
\textbf{Task Success and Execution Steps.}
For an evaluation set containing $N$ rollouts, let $s_i\in\{0,1\}$ indicate whether rollout $i$ succeeds and let $K_i$ denote its execution-step count.
Success Rate (SR) includes all rollouts, whereas the reported execution steps average only successful episodes, as shown in 
\textcolor{orange}{Eq.~(\ref{eq:xxxxx})}.
Failed rollouts therefore reduce SR but do not contribute to the reported mean completion steps.
Since each LIBERO suite contains the same number of evaluation rollouts, the four-suite mean SR also equals the success rate pooled across the four suites.
Overall execution steps are weighted by the number of successful episodes in each suite, rather than averaged equally across suite-level means.
\begin{equation}
\mathrm{SR}=\frac{100}{N}\sum_{i=1}^{N}s_i,
\qquad
\overline{K}=\frac{\sum_{i=1}^{N}s_iK_i}{\sum_{i=1}^{N}s_i}.
\label{eq:xxxxx}
\end{equation}

\textbf{Runtime Errors.}
We measure end-effector deviations from matched and aligned demonstration trajectories after each executed action chunk.
Translation error is the Euclidean distance between the executed and reference positions, and rotation error is the shortest angular distance between their orientations.
The reported units are millimeters and degrees, respectively.
Only successful episodes contribute to the task-evaluation error statistics, so the reported errors characterize deviations during task completion and are interpreted together with SR.
For LIBERO, the overall error is the arithmetic mean of the four suite-level error means, giving each suite equal weight.
The executed chunk contains five actions for $\pi_{0.5}$ and one action for GR00T-N1.5 on LIBERO.
These execution checkpoints are distinct from the internal integration steps used to generate actions.

\textbf{Inference Efficiency.}
Step speedup measures model inference time per policy query, while Rollout speedup measures cumulative model inference time over an episode.
If an episode contains $Q$ policy queries with inference times $\ell_1,\ldots,\ell_Q$, its cumulative inference time is $\sum_{t=1}^{Q}\ell_t$.
Both speedups normalize the corresponding BF16 inference time by that of the evaluated method, so a value greater than one indicates faster inference.
Rollout speedup captures the combined effects of query latency and the number of policy queries required during execution.
Consequently, a method with faster individual queries can still have a rollout speedup below one if task execution requires sufficiently more queries.
The memory values in the LIBERO efficiency comparison represent static full-model weight footprints in GB.

\subsection{Runtime Errors Across LIBERO Suites}

\begin{wrapfigure}{r}{42mm}
    \centering
    \includegraphics[width=\linewidth]{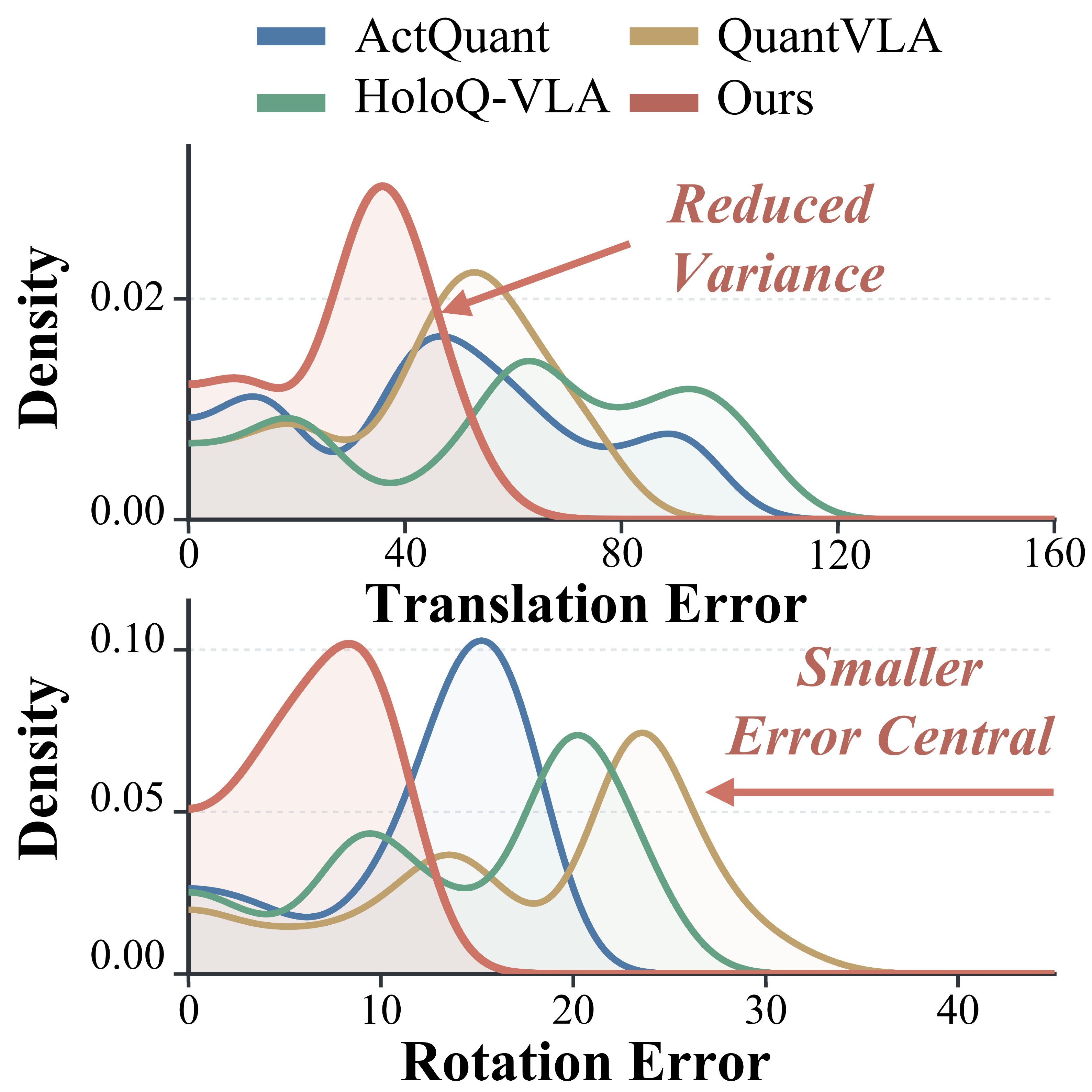}
    \captionsetup{position=bottom}
    \caption{Error Densities.}
    \label{fig:error_long}
\end{wrapfigure}

\textbf{Suite-Level Results.}
\textcolor{orange}{Tab.~\ref{tab:runtime_suite}} reports SR, translation error, rotation error, and execution steps for $\pi_{0.5}$ across all four LIBERO suites.
Among the methods listed, \textit{DyQ-VLA} achieves the lowest translation and rotation errors and the fewest execution steps in each suite.
On LIBERO-Long, both \textit{DyQ-VLA} and QuantVLA achieve 96.0\% SR, while translation error decreases from 91.98 to 62.45 mm, rotation error from 26.60 to 14.04 degrees, and mean execution steps from 224.5 to 163.3.
Thus, the same final success rate can accompany substantially different trajectory deviations and execution lengths.

\textbf{Runtime Error Distributions.}
\textcolor{orange}{Fig.~\ref{fig:error_long}} compares the translation and rotation error distributions on LIBERO-Long.
\textit{DyQ-VLA} exhibits a more concentrated translation-error distribution and a rotation-error distribution centered at smaller values.
These distributions complement the suite-level averages by showing the dispersion and location of the observed errors.
\par


\section{Real-World Evaluation Details}
\label{app:real_world_details}
We evaluate \textit{DyQ-VLA} on real-world tabletop manipulation tasks to examine whether its inference efficiency translates into faster task completion while preserving task performance.
This section describes the robotic platform, task coverage, policy configurations, and evaluation protocol.

\subsection{Robotic Platform And Workspace}
\label{apdx:RealWorldExpr-Platform}
\begin{figure}[t]
    \centering
    \includegraphics[width=\linewidth]{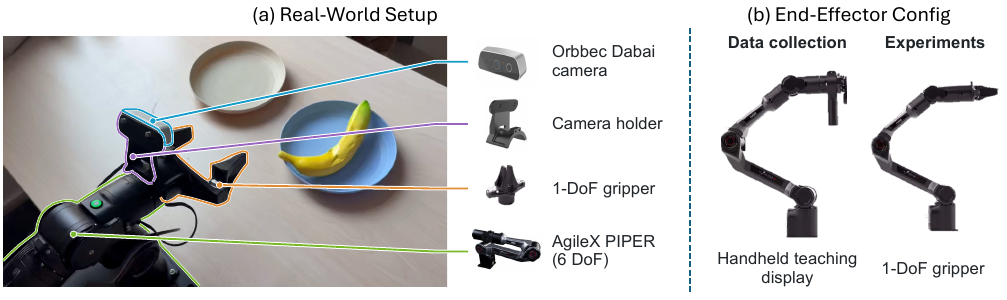}
    \caption{\textbf{Real-World Platform And End-Effector Configurations.}
    (a) Tabletop Workspace And Hardware Components.
    (b) Configurations for Demonstration Collection And Task Execution.}
    \label{fig:app_real_set}
\end{figure}

\textbf{Hardware Setup.}
The experiments use an Agilex Piper-X robotic arm equipped with a one-degree-of-freedom gripper and an Orbbec Dabai camera.
\textcolor{orange}{Fig.~\ref{fig:app_real_set}~(a)} identifies the arm, gripper, camera, and camera holder in the tabletop workspace.
The manipulation objects include fruit models, plates, cups, and a case with a lid.
These objects support both individual placement and insertion operations and tasks that combine several successive operations.

\subsection{Task Design}
\label{apdx:RealWorldExpr-Tasks}
\textbf{Evaluation Setting.}
Existing robotic resources support different aspects of policy learning and evaluation.
Open X-Embodiment~\citep{OXE} provides demonstrations from multiple robot embodiments, RoboCasa~\citep{robocasa} provides simulated kitchen manipulation environments and demonstration data, and SimplerEnv~\citep{simplerenv} evaluates policies in simulation setups corresponding to real robots.
\textcolor{orange}{Tab.~\ref{tab:robotic_settings}} summarizes these roles.
Our tabletop platform supports direct evaluation of task success and task completion time during physical execution.

\begin{table}[t]
\centering
\caption{Overview of Robotic Data And Evaluation Settings.}
\label{tab:robotic_settings}
\begingroup
\small
\setlength{\tabcolsep}{4pt}
\renewcommand{\arraystretch}{1.15}
\begin{tabular}{@{}
  >{\raggedright\arraybackslash}p{0.15\linewidth}
  >{\raggedright\arraybackslash}p{0.13\linewidth}
  >{\raggedright\arraybackslash}p{0.31\linewidth}
  >{\raggedright\arraybackslash}p{\dimexpr0.41\linewidth-24pt\relax}
@{}}
\toprule
\textbf{Resource} & \textbf{Setting} & \textbf{Scope} & \textbf{Primary Role} \\
\midrule
OXE
& Real-world
& Demonstrations across multiple robot embodiments
& Robot learning data \\
\addlinespace[3pt]
RoboCasa
& Simulation
& Household manipulation in kitchen environments
& Task environments and demonstration data \\
\addlinespace[3pt]
SimplerEnv
& Simulation
& Manipulation setups corresponding to real robots
& Policy evaluation in simulation \\
\addlinespace[3pt]
Our setup
& Real-world
& Six tabletop tasks on a single robotic platform
& Task success and completion-time evaluation \\
\bottomrule
\end{tabular}
\endgroup
\end{table}

\begin{figure}[t]
    \centering
    \includegraphics[width=\linewidth]{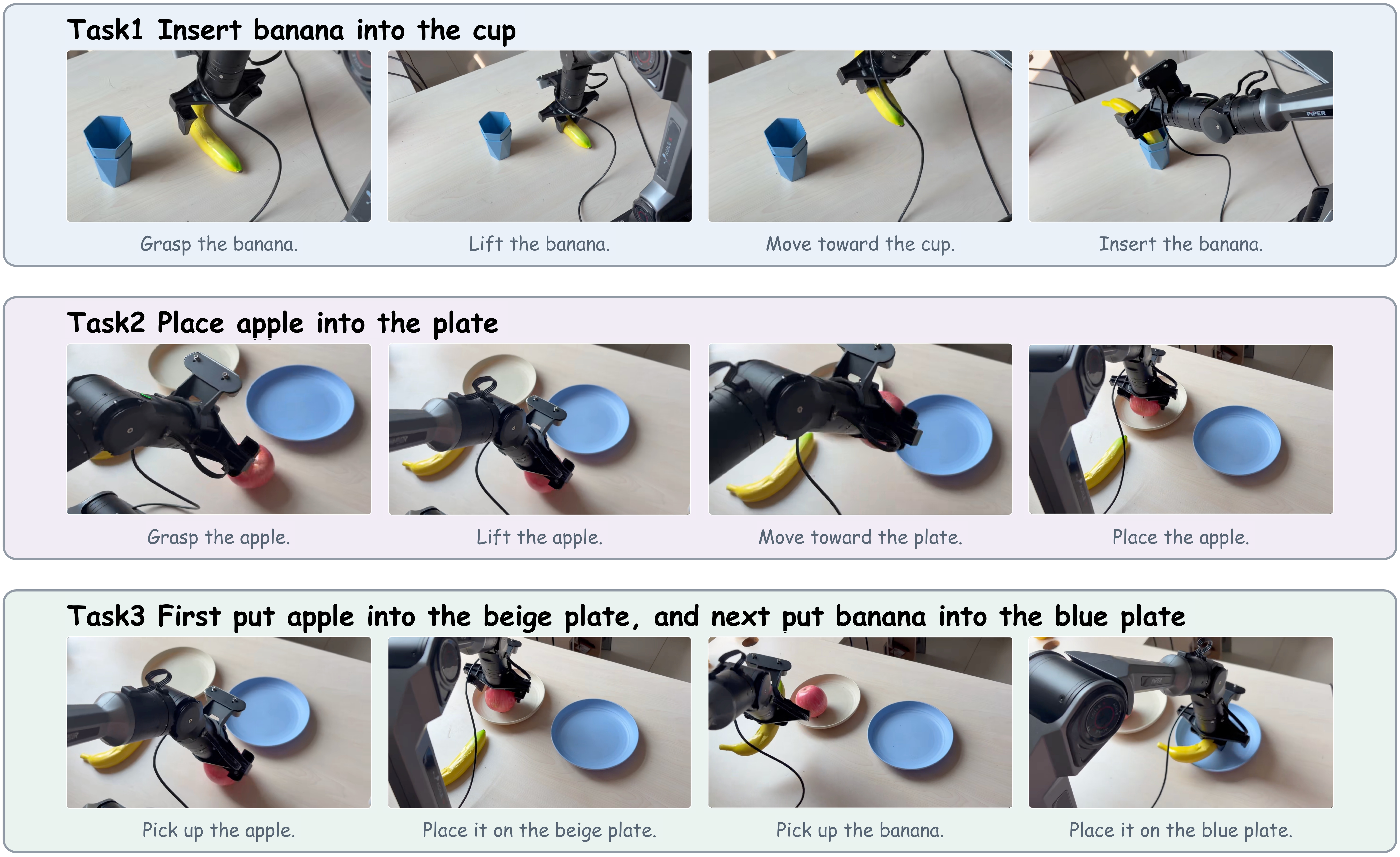}
    \caption{\textbf{Representative Real-World Task Sequences.}
    The examples show banana insertion, apple placement, and ordered placement of the apple and banana on their respective plates.}
    \label{fig:app_real}
\end{figure}

\textbf{Task Coverage.}
We consider four short-horizon tasks and two multi-stage tasks.
The short-horizon tasks involve placing a banana or an apple on a plate, inserting one cup into another, and inserting a banana into a cup.
The multi-stage tasks extend this coverage to orange packing followed by lid closure and placement of two objects in a specified order.
\textcolor{orange}{Tab.~\ref{tab:real_world_tasks}} lists the objective of each task, and \textcolor{orange}{Fig.~\ref{fig:app_real}} illustrates representative execution sequences.

\begin{table}[t]
\centering
\caption{Real-World Tasks And Objectives.}
\label{tab:real_world_tasks}
\begingroup
\small
\setlength{\tabcolsep}{5pt}
\renewcommand{\arraystretch}{1.15}
\begin{tabular}{@{}
  >{\raggedright\arraybackslash}p{0.19\linewidth}
  >{\raggedright\arraybackslash}p{0.23\linewidth}
  >{\raggedright\arraybackslash}p{\dimexpr0.58\linewidth-20pt\relax}
@{}}
\toprule
\textbf{Task Group} & \textbf{Task} & \textbf{Objective} \\
\midrule
Short-horizon
& Banana plate & Place the banana on the beige plate. \\
\addlinespace[3pt]
& Cup insertion & Insert one cup into the other cup. \\
\addlinespace[3pt]
& Banana cup & Insert the banana into the cup. \\
\addlinespace[3pt]
& Apple plate & Place the apple on the plate. \\
\midrule
Multi-stage
& Orange packing & Place the orange in the case and close the lid. \\
\addlinespace[3pt]
& Ordered placement & First place the apple on the beige plate, then place the banana on the blue plate. \\
\bottomrule
\end{tabular}
\par\vspace{3pt}
\begin{minipage}{\linewidth}
\small
$^{*}$Note: Each method is evaluated in ten trials per task.
Performance results are reported in \textcolor{orange}{Tab.~\ref{tab:real_world}}.
\end{minipage}
\endgroup
\end{table}

\subsection{Data Collection And Policy Setup}
\label{apdx:RealWorldExpr-Policy}
\textbf{Teaching And Execution Configurations.}
\textcolor{orange}{Fig.~\ref{fig:app_real_set}~(b)} distinguishes the two end-effector configurations used for data collection and experiments.
The handheld teaching configuration provides an interface for manually guided demonstrations, whereas the gripper configuration supports object manipulation during task execution.
This distinction separates demonstration acquisition from the autonomous executions used to evaluate the policy.

\textbf{Policy Backbone.}
The real-world evaluation uses $\pi_{0.5}$~\citep{pi_0.5}, with its BF16 execution serving as the reference for task performance and completion time.
All comparisons use this backbone under the precision configurations described below.

\subsection{Quantization And Online Execution}
\label{apdx:RealWorldExpr-Quantization}

\textbf{Precision Configurations.}
We compare \textit{DyQ-VLA} with QuantVLA~\citep{quantvla}, ActQuant~\citep{actquant}, QVLA~\citep{qvla}, and HoloQ-VLA~\citep{holoqvla}.
Their evaluated configurations are W4A8, W4A16, W4A16, and W4A4, respectively, where W and A denote weight and activation precision.
\textit{DyQ-VLA} maintains 4-bit weights and selects activation precision from $\{2,4,8,16\}$ bits during execution.
Thus, the comparisons cover fixed precision configurations and the dynamic activation precision used by \textit{DyQ-VLA}.

\textbf{Calibration Protocol.}
All quantized methods use the same 60 LIBERO calibration trajectories.
For \textit{DyQ-VLA}, offline calibration records the error levels and inference latency of the candidate activation configurations.
These statistics support precision selection through table lookup and constraint evaluation during execution.

\textbf{Runtime Error Control.}
Before each policy query, \textit{DyQ-VLA} uses estimated response sensitivity and historical rotational residuals to select the lowest-latency configuration satisfying the error constraints.
Selection uses precomputed calibration statistics rather than running every candidate configuration online.
The rotational components of the actions to be executed are then adjusted using residual-based compensation.
Precision-specific operators and fused execution support this dynamic inference path.
Together, these components connect numerical precision selection with action generation and compensation during execution.

\subsection{Evaluation Protocol}
\label{apdx:RealWorldExpr-Evaluation}
\textbf{Trials And Task Success.}
Each method is evaluated in ten trials per task, giving 60 trials across the six tasks.
For each task, we report the successful-trial count out of ten, and compute the overall success rate by dividing the total number of successful trials by 60.
Because every task has the same number of trials, this overall rate also equals the arithmetic mean of the six task-level success rates.
Both successful and unsuccessful trials contribute to this metric; the task objectives and their grouping are listed in \textcolor{orange}{Tab.~\ref{tab:real_world_tasks}}.
\textit{DyQ-VLA} succeeds in 44 of 60 trials (73.3\%), compared with 45 of 60 trials (75.0\%) for BF16.

\textbf{Task Completion Time.}
We report end-to-end task completion speedup relative to BF16 as Task Spd.
This metric measures the time required for physical task completion.
Step and Rollout speedups instead measure model inference time per query and cumulative model inference time per episode, respectively.
\textit{DyQ-VLA} achieves a $1.59\times$ task completion speedup over BF16, compared with $1.16\times$ for QuantVLA, the fastest quantized baseline.

\end{document}